\documentclass[conference]{IEEEtran}
\IEEEoverridecommandlockouts

\def\BibTeX{{\rm B\kern-.05em{\sc i\kern-.025em b}\kern-.08em
    T\kern-.1667em\lower.7ex\hbox{E}\kern-.125emX}}
    
\usepackage{graphicx}

\usepackage{float,epstopdf}
\usepackage{bbm}

\usepackage{microtype}

\usepackage{subcaption}
\usepackage{booktabs}
\usepackage{glossaries}
\usepackage{here}

\usepackage{amsmath}
\usepackage{amssymb}
\usepackage{mathtools}
\usepackage{amsthm}
\usepackage{dsfont}
\usepackage{multicol}
\usepackage{multirow} 
\usepackage{amsfonts} 
\usepackage{mathrsfs}
\usepackage{fancyhdr}
\usepackage[amssymb, thickqspace]{SIunits}
\usepackage{enumitem}

\usepackage{pgfplotstable}
\usepackage{arydshln}
\usepackage{lipsum}		

\usepackage{natbib}

\setcitestyle{numbers}
\setcitestyle{square}

\usepackage{cases}

\usepackage{algorithm}
\usepackage{algorithmic}

\usepackage{url}

\usepackage[textsize=tiny]{todonotes}

\newcounter{row}
\makeatletter
\@addtoreset{subfigure}{row}
\makeatother

\newlength{\tempheight}
\newlength{\tempwidth}

\newcommand{\rowname}[1]
{\rotatebox{90}{\makebox[\tempheight][c]{\textbf{#1}}}}

\newcommand{\columnname}[1]
{\makebox[\tempwidth][c]{\textbf{#1}}}
\makeatletter
\def\munderbar#1{\underline{\sbox\tw@{$#1$}\dp\tw@\z@\box\tw@}}
\makeatother

\newcommand{\be}{\begin{equation}}
\newcommand{\ee}{\end{equation}}
\newcommand{\bea}{\begin{equation*}\begin{aligned}}
\newcommand{\eea}{\end{aligned}\end{equation*}}

\newcommand{\R}{\mathbb{R}}

\newcommand{\mc}{\mathcal}

\newcommand{\diff}{\mathrm{d}}
\newcommand{\softmax}{\mathrm{softmax}}
\newcommand{\dist}{\mathrm{dist}}
\newcommand{\Emb}{\mathrm{Emb}}

\newacronym{wsn}{WSN}{wireless sensor network}
\newacronym{wrsn}{WRSN}{wireless rechargeable sensor network}
\newacronym{sn}{SN}{sensor node}
\newacronym{rn}{RN}{relay node}
\newacronym{bs}{BS}{base station}
\newacronym{pf}{PF}{Pareto front}
\newacronym{mst}{MST}{minimum spanning tree algorithm}
\newacronym{dem}{DEM}{digital elevation model}
\newacronym{drl}{DRL}{deep reinforcement learning}
\newacronym{iot}{IoT}{Internet of Things}
\newacronym{ai}{AI}{Artificial Intelligent}
\newacronym{mc}{MC}{mobile charger}
\newacronym{njnp}{NJNP}{Nearest-Job-Next with Preemption}
\newacronym{inma}{INMA}{Invalid Node Minimized Algorithm}
\newacronym{drltcc}{DRL-TCC}{deep reinforcement learning approach for target coverage and connectivity problem}
\newacronym{qos}{QoS}{Quality of Service}
\newacronym{ia}{IA}{intelligent agent}
\newacronym{rl}{RL}{reinforcement learning}
\newacronym{dl}{DL}{deep learning}
\newacronym{mdp}{MDP}{Markov decision process}

\newglossary[nlg]{notation}{not}{ntn}{Notation}

\newglossaryentry{not:eth}{
    name=\ensuremath{\tilde{E}_{td}},
    description={energy requesting threshold},
    type=notation}
    
\newglossaryentry{not:bs}{
    name=\ensuremath{p^{\mathrm{BS}}},
    description={base station},
    type=notation}
    
\newglossaryentry{not:depot}{
    name=\ensuremath{p^{\mathrm{D}}},
    description={depot},
    type=notation}
    
\newglossaryentry{not:snset}{
    name=\ensuremath{\mathcal{P}},
    description={a set of deployed sensors},
    type=notation}

\newglossaryentry{not:num:sn}{
    name=\ensuremath{n},
    description={number of deployed sensors},
    type=notation}
    
\newglossaryentry{not:sn}{
    name=\ensuremath{p},
    description={a sensor},
    type=notation}
    
\newglossaryentry{not:tgset}{
    name=\ensuremath{\mathcal{Q}},
    description={a set of critical targets},
    type=notation}
    
\newglossaryentry{not:num:tg}{
    name=\ensuremath{m},
    description={number of critical targets},
    type=notation}
    
\newglossaryentry{not:tg}{
    name=\ensuremath{q},
    description={a critical target},
    type=notation}

\newglossaryentry{not:r:s}{
    name=\ensuremath{r_s},
    description={sensing range},
    type=notation}

\newglossaryentry{not:r:c}{
    name=\ensuremath{r_c},
    description={communication range},
    type=notation}
    
\newglossaryentry{not:mu}{
    name=\ensuremath{\mu},
    description={charging rate},
    type=notation}
    
\newglossaryentry{not:v}{
    name=\ensuremath{\nu},
    description={velocity of the MC},
    type=notation}
    
\newglossaryentry{not:c:mc}{
    name=\ensuremath{B^{\mathrm{MC}}},
    description={battery capacity of the MC},
    type=notation}

\newglossaryentry{not:c:sn}{
    name=\ensuremath{B^{\mathrm{SN}}},
    description={battery capacity of a sensor},
    type=notation}
    
\newglossaryentry{not:e:move}{
    name=\ensuremath{\omega_{\mathrm{move}}},
    description={ECR of the MC for traveling},
    type=notation}
    
\newglossaryentry{not:ecr}{
    name=\ensuremath{\omega},
    description={energy consumption rate},
    type=notation}
    
\newglossaryentry{not:traj}{
    name=\ensuremath{\tau},
    description={charging trajectory},
    type=notation}
    
\newglossaryentry{not:ac}{
    name=\ensuremath{a},
    description={charging action},
    type=notation}
    
\newglossaryentry{not:re:sn}{
    name=\ensuremath{e},
    description={residual energy of a sensor},
    type=notation}
    
\newglossaryentry{not:re:mc}{
    name=\ensuremath{e^{\mathrm{MC}}},
    description={residual energy of the MC},
    type=notation}
    
\newglossaryentry{not:acset}{
    name=\ensuremath{\mathcal{A}},
    description={a set of legal actions},
    type=notation}

\newglossaryentry{not:policy}{
    name=\ensuremath{\pi},
    description={a policy},
    type=notation}
    
\newglossaryentry{not:state:space}{
    name=\ensuremath{\mathcal{S}},
    description={a state space},
    type=notation}

\newglossaryentry{not:state}{
    name=\ensuremath{x},
    description={a state},
    type=notation}
    
\newglossaryentry{not:state:mc}{
    name=\ensuremath{x^{MC}},
    description={state of the mobile charger},
    type=notation}

\newglossaryentry{not:state:depot}{
    name=\ensuremath{x^{D}},
    description={state of the depot},
    type=notation}
    
\newglossaryentry{not:state:sn}{
    name=\ensuremath{x^{SN}},
    description={state of a sensor},
    type=notation}

\newglossaryentry{not:transition}{
    name=\ensuremath{T},
    description={a transition model},
    type=notation}

\newglossaryentry{not:reward:func}{
    name=\ensuremath{R},
    description={a reward function},
    type=notation}
    
\newglossaryentry{not:mdp}{
    name=\ensuremath{\mathcal{M}},
    description={Markov decision process},
    type=notation}
    
\newglossaryentry{not:gamma}{
    name=\ensuremath{\gamma},
    description={a discount factor (discount rate)},
    type=notation}
\usepackage[normalem]{ulem}
\makeglossaries 

\begin{document}


\title{A Deep Reinforcement Learning-based Adaptive Charging Policy for WRSNs}

\author{\IEEEauthorblockN{Ngoc Bui\IEEEauthorrefmark{1}\IEEEauthorrefmark{2}, Phi Le Nguyen\IEEEauthorrefmark{1}, Viet Anh Nguyen\IEEEauthorrefmark{2}, Phan Thuan Do\IEEEauthorrefmark{1}}
\IEEEauthorblockA{
\IEEEauthorrefmark{1}\textit{School of Information and Communication Technology, Hanoi University of Science and Technology}\\
\IEEEauthorrefmark{2}\textit{VinAI Research}\\
Hanoi, Vietnam \\
Email: \IEEEauthorrefmark{1}\{ngoc.bh212155m@sis, lenp@soict, thuandp@soict\}.hust.edu.vn, \IEEEauthorrefmark{2}v.anhnv81@vinai.io}
}

\maketitle

\begin{abstract}
    Wireless sensor networks consist of randomly distributed sensor nodes for monitoring targets or areas of interest. Maintaining the network for continuous surveillance is a challenge due to the limited battery capacity in each sensor. Wireless power transfer technology is emerging as a reliable solution for energizing the sensors by deploying a mobile charger (MC) to recharge the sensor. However, designing an optimal charging path for the MC is challenging because of uncertainties arising in the networks. The energy consumption rate of the sensors may fluctuate significantly due to unpredictable changes in the network topology, such as node failures. These changes also lead to shifts in the importance of each sensor, which are often assumed to be the same in existing works. We address these challenges in this paper by proposing a novel adaptive charging scheme using a deep reinforcement learning (DRL) approach. Specifically, we endow the MC with a charging policy that determines the next sensor to charge conditioning on the current state of the network. We then use a deep neural network to parametrize this charging policy, which will be trained by reinforcement learning techniques. Our model can adapt to spontaneous changes in the network topology. The empirical results show that the proposed algorithm outperforms the existing on-demand algorithms by a significant margin.
\end{abstract}

\section{Introduction}
In recent decades, the Internet of Things (IoT) has been flourishing thanks to the advancements in electronic, communication, and computing technologies. Wireless sensor networks (WSNs), which originated in military use, are becoming one of the main building blocks enabling the ubiquity of the IoT in many civilian applications, for example, home automation~\citep{pirbhulal2016novel}, air/earthquake monitoring~\citep{kingsy2019comprehensive, alphonsa2016earthquake}, smart agriculture~\citep{sanjeevi2020precision}, health monitoring~\citep{abdulkarem2020wireless, gardavsevic2020emerging}, smart cities~\citep{csaji2017wireless}. Generally, WSNs can be defined as self-configured and infrastructure-less wireless networks comprised of multiple spatially dispersed and dedicated sensors to monitor specific targets or areas of interest. The sensing data will be cooperatively transmitted through the wireless network to a base station (BS), also known as a \textit{sink}, where the data can be observed and analyzed. 

The real-world deployment of WSNs must fulfill many quality-of-service (QoS) requirements, in which coverage and connectivity are often considered two paramount factors~\cite{tripathi2018coverage}. The coverage specifies how well the areas or targets of interest are monitored by the sensors, while the connectivity relates to the ability to transmit the sensing data from the sensors to the BS. Ensuring the coverage and connectivity is critical since, in many applications, the network is required to monitor and analyze the targets or areas continuously~\cite{zhao2008lifetime}.

However, maintaining the network for continuous surveillance is an enormous challenge due to the energy restriction of the sensors--that the sensors are often equipped with low-cost, low-power batteries~\cite{akyildiz2002wireless}. When the battery is fully consumed, a sensor is no longer able to monitor the targets or relay the data; thus, the network may become fragmented and the data from some parts of the sensing field may no longer be extracted. Furthermore, sensor networks are often implemented on a wide scale in potentially hazardous terrain that is difficult for people to access (e.g., battlefields, underground). In such circumstances, it is difficult to replace the sensors' batteries.

Traditional approaches focus on conserving energy via optimizing sensor functioning, such as data reduction~\citep{goyal2019data}, sleep/wakeup schemes~\citep{haimour2019energy}, and energy-efficient routing~\citep{raj2019qos}. However, this approach is only able to extend the sensors' lifetime for a certain amount of time. The battery will eventually be exhausted if there is no external source supplying the sensors. An alternative solution is to deploy an energy harvester or scavenger inside each sensor to convert energy from an external source (e.g., solar, thermal, wind)~\citep{adu2018energy}. Nevertheless, this technique dramatically depends on an ambient source that is usually unstable and uncontrollable.


Recent progress in wireless power transmission technology based on electromagnetic waves~\citep{kurs2007wireless, lu2015wireless} has given rise to a novel scheme, namely \textit{wireless rechargeable sensor networks} (WRSNs)~\cite{he2012energy}, for energizing the sensors. The idea is to employ a (or multi-) mobile charger (MC), which is equipped with a high-capacity battery and a transmission coil, to travel around the sensing field and charge the sensors wirelessly. Here, the sensors have a receiver that helps the sensors receives energy from the MC through electromagnetic waves. Unlike the energy harvesting techniques, this scheme offers agile, controllable, and reliable energy replenishment, thus enabling genuinely sustainable operations for the sensor networks.

In practice, the charging strategy of the MC(s) has a considerable impact on the performance of a WRSN (i.e., sensor lifetime). Numerous research has been conducted to design charging strategies for the MC, which can be classified into two categories: \textit{periodic} charging and \textit{on-demand} charging. In the periodic charging scheme, the energy consumption rate of each sensor is supposed to be constant and known in advance. Thereby, an optimal charging trajectory could be planned before the running phase. The MC then travels along the pre-optimized charging trajectory to recharge nodes in a periodic and deterministic manner~\cite{lyu2019periodic, jiang2017joint, ma2018charging, xu2019minimizing}. However, the sensors' energy consumption profiles often fluctuate greatly due to the close interaction with the surrounding environment~\cite{he2013demand}. Furthermore, WSNs are very dynamic--that a node failure might change the routing paths of many packets, causing turbulence in the average power consumption of many nodes. As a result, the pre-optimized charging trajectory may become inefficient.



To overcome these issues, the on-demand charging scheme instead requires the sensors to send a charging request to the MC when their residual energy falls below a predetermined threshold. The MC will maintain a pool of charging requests and then choose the next charging destinations based on the current service pool. He et al.~\citep{he2013demand} introduced a simple heuristic algorithm, namely \gls{njnp}, that chooses the next charging destination according to the spatially closest sensor in the queue. Several following works~\citep{fu2015esync, lin2017tsca, lin2019double, zhu2018adaptive, kaswan2018efficient} improved this heuristic by introducing double warning thresholds or using meta-heuristic algorithms. Recent works also leveraged (deep) Q-learning to learn on-demand charging policy~\citep{la2020q, cao2021deep}.

\begin{figure}[bt]
    \centering
    \includegraphics[width=0.8\linewidth]{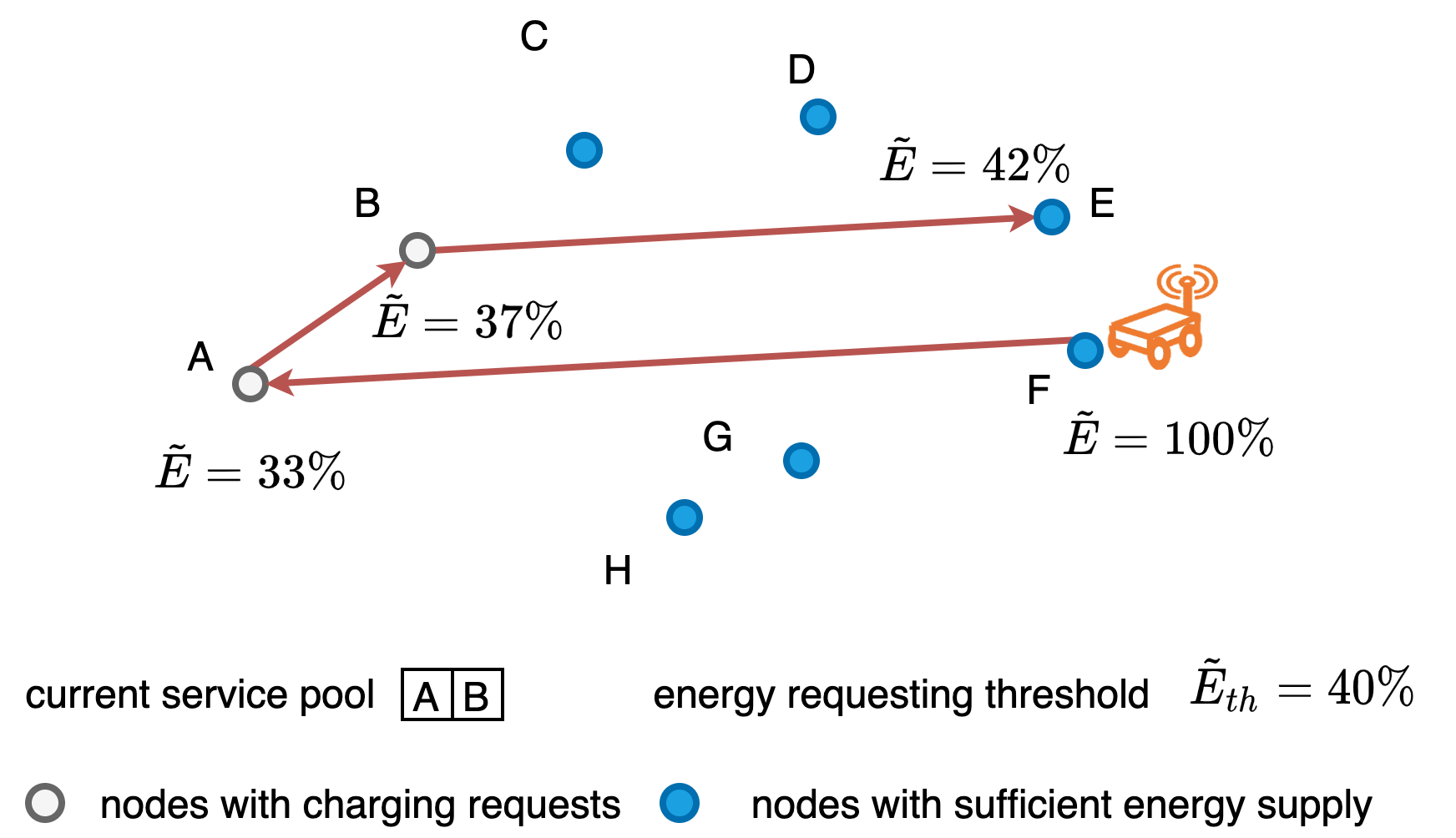}
    \caption{The drawback of on-demand charging scheme. Node $E$ sends a charging request right after the \glsentrytext{mc} decides to charge node $A$ next.} 
    \label{fig:threshold_example}
    \vspace{-5mm}
\end{figure}

Despite the promising results, a common drawback of the on-demand algorithms is the dependence on the chosen threshold for the charging requests. Specifically, if the charging threshold is set too high, the sensors will send the charging requests too frequently, causing the MC to become overloaded and degrading the efficiency of the charging algorithms~\cite{zhu2018adaptive}. A small charging threshold, on the other hand, may cause sensors to submit requests too late, and the MC may not come in time to charge before the sensor runs out of energy. 
In the example shown in Fig. \ref{fig:threshold_example}, after fully charging node $F$, the \gls{mc} continuously chooses a requested node in the current service pool to charge (node $A$ or $B$). If node $E$ requests charging when the \gls{mc} has just decided to go to $A$ or $B$, the \gls{mc} must move back and forth to serve the requests.


In this article, we propose a novel scheme called \textit{adaptive charging} to address the drawbacks of the existing approaches. Specifically, we eliminate the energy threshold in the on-demand charging and endow the MC with the ability to choose any sensor to charge at any time depending on the current state of the network. The charging policy thus is a mapping between the network's state space and all possible actions which specifies what to do at a given state. We leverage a deep neural network (DNN) to approximate the strategy's mapping and use reinforcement learning (RL) techniques to train the model. The MC will act as an agent in the sensor network (environment), learning how to derive the optimum charging decisions by interacting with the environment. We will train the model with a sufficiently large number of network topologies in advance; thus, the trained model can be applied to any network topology without manually adjusting the parameters. 

\textbf{Contributions. } We consider in this paper the connected target coverage problem \cite{zhao2008lifetime}, under the setting of the \gls{wrsn} paradigm. Specifically, several critical targets are required to be continuously monitored by a number of randomly scattered sensors, which can be replenished wirelessly by an MC. Our objective is to design a charging strategy for the MC to maximize the time interval that all targets are continuously monitored and analyzed by the BS. The main contributions of this paper are as follows.

\begin{itemize}
    \item We propose an adaptive charging scheme by omitting the energy threshold for charging requests in the on-demand scheme and endowing the MC ability to charge any sensor at any time. This enables the MC to evaluate other charge options that may improve the cumulative network's lifespan instead of focusing on the service pool.
    \item We design a charging policy to prolong the network's lifetime using a DNN model, which can be trained by RL techniques. Our model is flexible--it can operate in a dynamic sensor network where the number of sensors might change due to node failures or deployments.
    \item We conduct the experiment showing the superiority of our model compared to existing on-demand methods in prolonging the network lifetime. We then discuss further the self-organizing capability of the proposed method.
\end{itemize}


The rest of the paper is unfolded as follows: In the next section, we describe the system model and the setup of the connected target coverage problem in WRSNs. Section~\ref{sec:model_arc} presents our learning model for the adaptive charging strategy of the MC. In Section~\ref{sec:expt}, we conduct experiments to demonstrate the efficiency of the proposed method.

\section{System Model and Problem Statement}\label{sec:problem}

\subsection{System Model}

\begin{figure}[tb]
    \centering
    \includegraphics[width=0.9\linewidth]{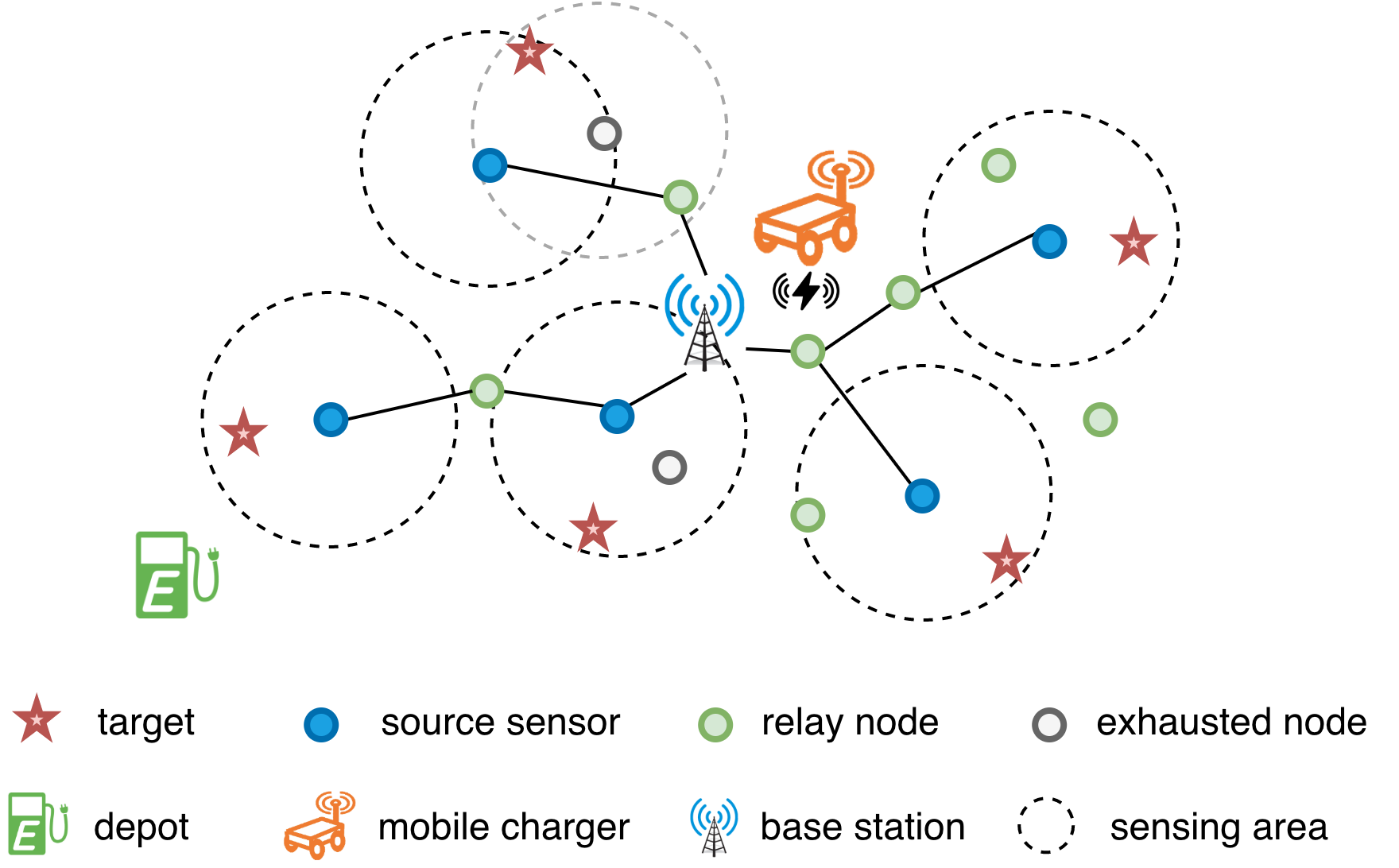}
    \caption{An illustration of a wireless rechargeable sensor network for target-covering.}
    \vspace{-5mm}
    \label{fig:network_model}
\end{figure}

Figure~\ref{fig:network_model} depicts our targeted wireless rechargeable sensor network, which comprises four main components: sensors, a base station, a depot, and a mobile charger.
Each sensor has a sensing unit, a processing unit, a transceiver unit, and a power unit.
The sensors are responsible for monitoring pre-specified targets and transmitting sensory data to the base station in a multi-hop paradigm.
A mobile charger is a device equipped with a wireless energy transfer module. It will travel around the network and wirelessly transmit power to the sensors.
When the mobile charger's energy is almost exhausted, it will return to the depot to charge.

Each sensor has a sensing area determined by its location and sensing range. Any target located in the sensing area of a sensor could be monitored. A sensor is called a \textit{source node} if it covers at least one target, while the sensors that do not cover any target may act as relay nodes forwarding the sensing data to the sink. A source sensor will perform the monitoring task and periodically generate sensing data. Here, we assume that all source sensors generate sensing data at the same rate, i.e., all sensors have the same sampling frequency, quantization, and coding scheme~\cite{zhao2008lifetime}. Thus, each source sensor generates a fixed amount of bits per unit of time. The sensing data gathered by source sensors is transmitted to the sink by multi-hop communication over other sensors. Two sensors can communicate with each other if the distance between them is less than a communication range. 



\subsection{Energy Consumption Model}


We use the same energy model as in \citep{Gawade2016}, which accounts for the dissipated energy of the sensors at both the receiver and transmitter during transmission while omitting the dissipated energy of the sensing and computing units. The energy dissipated by the transmitter for transmitting a $k$-bit packet to a distance $l$ is given by:
\begin{equation}
  \tilde{E_t}(k, l) = 
  \begin{dcases}
    k \epsilon_{elec} + k \epsilon_{fs} l^2, & \text{if } l \leq l_0 \leq r_c,\\
    k \epsilon_{elec} + k \epsilon_{mp} l^4, & \text{if } l_0 < l \leq r_c,\\
    \infty, & \text{if } r_c < l,
\end{dcases}
  \label{eq:transmit}
\end{equation}
where $l_0 = \sqrt{\frac{\epsilon_{fs}}{\epsilon_{mp}}}$ is the distance threshold for swapping amplification models and $r_c$ indicates the range within which a node can communicate. The element $\epsilon_{elec} = 50 nJ/bit$ is the energy dissipated per bit to run the transmitter or receiver circuit,; $\epsilon_{fs} = 10 pJ/bit/m^2$ and $\epsilon_{mp} = 0.0013 pJ/bit/m^4$ are the energy expenditures of transmitting one-bit data at a short and a long-distance respectively to achieve an acceptable bit error rate. These constants are set similar to~\citep{Wu2013}.

The energy consumption of the receiver to receive an $k$-bit packet is calculated as follows:
\begin{equation}
  \tilde{E}_r(k) = k \epsilon_{elec}.
  \label{eq:receive}
\end{equation}

The dissipated energy of a node receiving $\eta$ packets and transmitting them to the parent node is then calculated by the following formula:
\begin{equation}
  \tilde{E}_{\mathrm{SN}}(\eta, \zeta, l, k) = \eta \tilde{E}_r(k) + (\eta + \zeta)\tilde{E}_t(k, l),
  \label{eq:ecr_sn}
\end{equation}
where $\tilde{E}_t(k, l), \tilde{E}_r(k)$ are calculated as in Eq.~\eqref{eq:transmit} and~\eqref{eq:receive}, respectively. The parameter $\zeta$ is the number of targets surrounding the sensor while
$l$ is the transmission distance.

\subsection{Charging Model} \label{sec:charging_model}

In this paper, we only allow the MC to charge one sensor at a time and the sensor will be fully charged. A charging action of the MC consists of two phases: (1) moving into the vicinity of the sensor and (2) charging the sensor to full battery capacity. The time to charge a sensor is determined as follows:
\begin{equation}
    t^{\mathrm{ch}}_{i} = \frac{\gls{not:c:sn} - \gls{not:re:sn}_i}{\gls{not:mu} - \gls{not:ecr}_i},
\end{equation}
where $\mu$ is the charging rate, \gls{not:c:sn} and $\gls{not:re:sn}_i$ denote the battery capacity and the residual energy of the sensor, respectively. The parameter $\gls{not:ecr}[_i]$ is the energy consumption rate (ECR) of the sensor $i$ which will be estimated by the BS using the residual energy profile of the sensor. The residual energy profile of the sensor is the sequence of notification packets, i.e., the sensor records its residual energy and current time stamp and periodically sends it to the BS. This procedure is similar to the estimation of the dynamic energy consumption rate in~\cite{zhu2018adaptive}.

The dissipated energy of the MC for traveling $l$ units of distance and charging the sensor $i$ is calculated as:
\begin{equation}
    \tilde E_{\mathrm{MC}}(l, e_i, \omega_{\mathrm{move}}) = l \omega_{\mathrm{move}} + \mu t^{\mathrm{ch}}_i.
    \label{eq:ecr_mc}
\end{equation}
Here, similar to~\cite{cao2021deep}, we omit the energy dissipated into the environment during charging; thus the energy consumption of the MC is only composed of the energy transferred to sensor nodes and the energy consumed by traveling.



\subsection{Problem Statement}
]
Let us denote by $\gls{not:snset} = \{\gls{not:sn}[_1], \gls{not:sn}[_2], \ldots, \gls{not:sn}[_{\gls{not:num:sn}}]\}$ the set of $n$ deployed sensors and by $\gls{not:tgset} = \{\gls{not:tg}[_1], \gls{not:tg}[_2], \ldots, \gls{not:tg}[_{\gls{not:num:tg}}]\}$ the set of $m$ targets to be monitored. We also refer to \gls{not:bs} and \gls{not:depot} as the position of the BS and the depot, respectively. All nodes are deployed in a two-dimensional sensing area. All sensors have the same battery capacity~$\gls{not:c:sn}$, sensing range~$\gls{not:r:s}$, and communication range~$\gls{not:r:c}$. A target $q_k$ can be monitored by a sensor $p_i$ if $\dist(p_i, q_k) \leq r_s$, where $\dist(\cdot, \cdot)$ denotes the Euclidean distance. Meanwhile, two sensors $p_i$ and $p_j$ can communicate to each other if $\dist(p_i, p_j) \leq r_c$. Initially, we assume that each target is covered by at least one sensor and there exists at least one transmission route from each sensor toward the BS. When a sensor depletes its energy, it will deactivate itself and wait to be replenished.

A network's state is considered covered and connected if it satisfies the two following constraints: (1) \textit{coverage}, each target is covered by at least one source sensor, (2) \textit{connectivity}, from each source sensor to sink, there must exist at least one route traversing through only active sensors. We then define the \textit{network lifetime} as the time interval from when the network
starts till the target coverage or the connectivity is not satisfied.

The MC's parameters can be represented by four factors ${\langle\gls{not:c:mc}, \gls{not:v}, \gls{not:mu}, \gls{not:e:move}\rangle}$, where \gls{not:c:mc} is  the battery capacity, \gls{not:v} is the traveling speed, \gls{not:e:move} denotes the energy consumption rate of the MC per one unit distance, and \gls{not:mu} accounts for both charging rates from the MC to a sensor and from the depot to the MC. For the sake of simplicity, we assume that the velocity of the MC is constant and the MC can travel freely inside the sensor field without obstacles. 

The charging path of the MC is a sequence of charging locations which are the positions of the sensors or the depot. If the MC travels to a sensor, the MC will charge it to full capacity. If the MC goes back to the depot, the MC will recharge its own battery fully. Our objective is to determine a charging path that maximizes the lifetime of the sensor network so that every target is monitored and analyzed continuously by the BS.


Notice that our scheme does not require the sensors to send the charging requests to the MC. We endow the \gls{mc} the ability to choose any charging destination at any time based on the network's state and its own status. Thus we do not need to predetermine the energy threshold for charging requests like the existing on-demand charging schemes~\cite{he2013demand, zhu2018adaptive, cao2021deep}. 

\section{A DRL-based Adaptive Charging Scheme} \label{sec:scheme}



\begin{figure}
    \centering
    \includegraphics[width=0.8\linewidth]{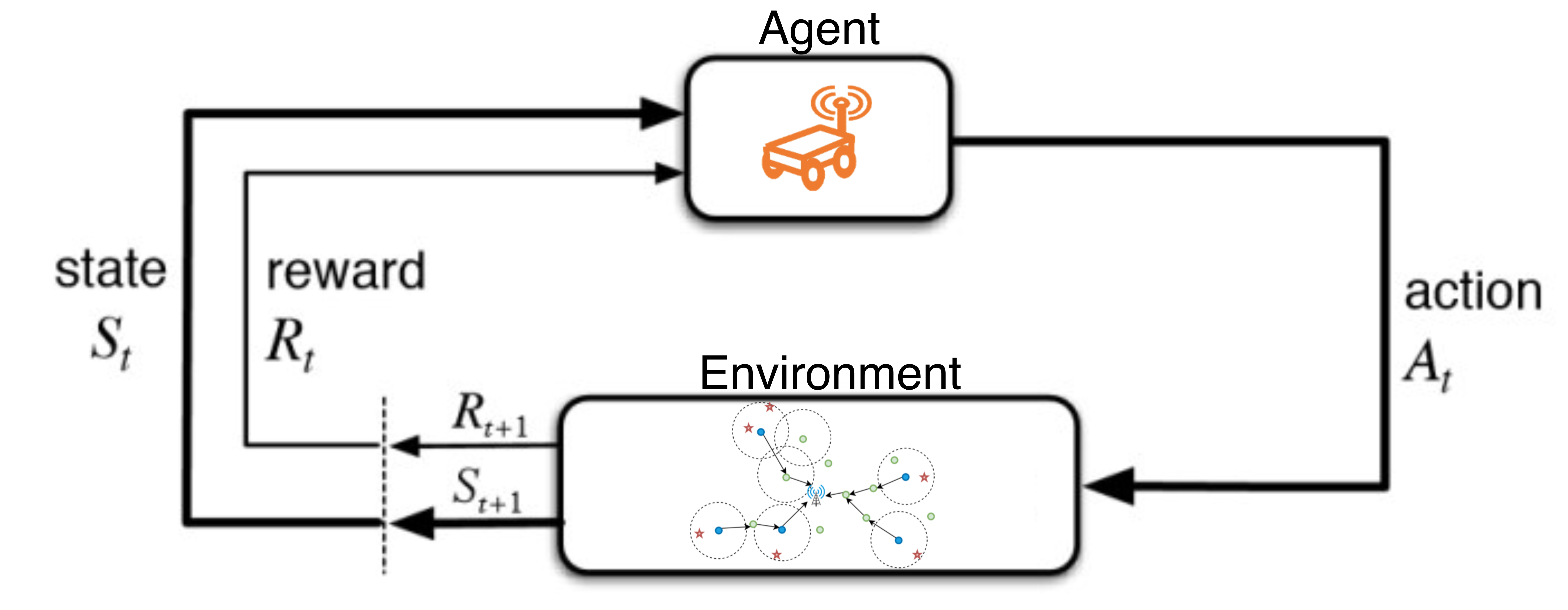}
    \caption{Learning model of a reinforcement learning system.}
    \label{fig:rl}
    \vspace{-5mm}
\end{figure}

The goal of this paper is to design a charging strategy for the MC that can adapt to the uncertainties arising in the sensor network. To this end, we propose a novel adaptive charging scheme based on deep reinforcement learning. The MC will act as an intelligent agent interacting with the sensor network by moving around to replenish sensors or itself~(Fig.~\ref{fig:rl}). We model the policy using a deep neural network that takes a network's state as the input and outputs the probabilities of taking charging decisions. The model is then trained by interacting with simulated environments and adjusting the policy to the return from the environments. 



\subsection{Formulation of the DRL Framework}

In the following, we formulate our problem under the DRL framework. The mathematical model of a DRL model typically consists of four items, namely a state space $\gls{not:state:space}$, an actions space $\gls{not:acset}$, a transition model $\gls{not:transition}$, and a reward function $\gls{not:reward:func}$.



\textbf{State. } 
The state information represents the status of the network and the MC that helps the MC choose the next charging locations. We divide the useful information into two groups: \textit{static} and \textit{dynamic} elements. The static elements contain prescribed information related to the properties of the mobile charger, the depot, or the sensors such as the position of sensors, battery capacity, and the number of targets covered by a sensor. The dynamic elements include temporal information such as the residual energy of devices, the current position of the \gls{mc}, and the estimated energy consumption rate of each sensor. Formally, we define a state $\gls{not:state} \in \gls{not:state:space}$ as a tuple of $(\gls{not:state:mc}, \gls{not:state:depot}, \bar{x}^{SN})$, where $x^{MC}$ is a tuple of the static and dynamic information of the mobile charger, $x^D$ contains the location of the depot in the sensor field, and $\bar{x}^{SN} = \{\gls{not:state:sn}[_i], i=1,\ldots, n\}$ is a sequence of tuples containing static and dynamic information of each sensor. The detail of the state information is shown in Table~\ref{tab:state_inf}. 
\begin{table}[tb!]
  \centering
  \caption{State information. The notation S and D indicate static and dynamic information, respectively.}
  \label{tab:state_inf}
  \footnotesize
  \begin{tabular}{llll}
    \toprule
     State & Parameter & Type & Comment \\
    \midrule
    \multirow{6}{*}{$x^{\mathrm{MC}}$}&\gls{not:c:mc} & S & \glsdesc*{not:c:mc} \\
    &\gls{not:v} & S & \glsdesc*{not:v} \\
    &\gls{not:mu} & S & \glsdesc*{not:mu} \\
    &\gls{not:e:move} & S & ECR for traveling \\
    & $p^{\mathrm{MC}}$ & D & current position \\
    & $e^{\mathrm{MC}}$ & D & residual energy power \\
    \midrule
    \multirow{5}{*}{$x^{\mathrm{SN}}_i$} & \gls{not:c:sn} & S & \glsdesc*{not:c:sn} \\
    & $p_i$ & S & position of sensor $i$ \\
    & $\xi_i$ & S & no.~of targets covered by sensor $i$ \\
    & $e_i$ & D & residual energy of sensor $i$ \\
    & $\omega_i$ & D & ECR of sensor $i$ \\
    \midrule
    $x^D$ & $p^{\mathrm{D}}$ & S & position of the depot \\
    \bottomrule
  \end{tabular}
  \vspace{-5mm}
\end{table} 

\textbf{Action. } We define $n+1$ actions corresponding to $n+1$ charging destinations (a depot and $n$ sensors). An action $a_t$ made at time $t$ is an integer number $ \gls{not:ac}[_t] \in \gls{not:acset} = \{0, 1, 2, ..., \gls{not:num:sn}\}$, where we denote $\gls{not:ac}[_t] = i$, $i > 0$ with regard to the integral index of the sensors, and $\gls{not:ac}[_t] = 0$ corresponds to going back to the depot and recharging itself.

\textbf{Transition. } We simulate the network environment and return the network's state at the end of doing the action $a_t$ as the next state $s_{t+1}$. The dissipated energy of the sensors and the MC will be computed following Eq.~\eqref{eq:ecr_sn} and~\eqref{eq:ecr_mc}. It is worth noting that the MC is only able to observe the returned state $s_{t+1}$, not the underlying process of the simulation.

\textbf{Reward. } We define the reward $R(\gls{not:state}_t, \gls{not:ac}_t)$ of doing an action $\gls{not:ac}_t$ at state $\gls{not:state}_t$ as the period of time doing the action which includes both traveling and charging time. If either the coverage or connectivity conditions are not met, the simulation environment will be terminated immediately and return the reward till the network downtime. Thus, cumulative reward over all actions coincides with the lifetime of the network.


 
A charging trajectory can be represented as a sequence of the network's states and the \gls{mc}'s actions $\gls{not:traj} = \{\gls{not:state}[_0], \gls{not:ac}[_0],\ldots, \gls{not:state}[_{T-1}],\gls{not:ac}[_{T-1}], \gls{not:state}[_T]\}$, $\gls{not:ac}[_t] \in \gls{not:acset}$, $\gls{not:state}[_t] \in \gls{not:state:space}$. 
The $\gamma$-discounted lifetime of the trajectory $\tau$ can be computed as:
\begin{equation}
    G(\gls{not:traj}) = \sum^{T}_{t=0} \gls{not:gamma}^{t} R(x_t, a_t).
\end{equation} 
A stochastic policy $\gls{not:policy}(\gls{not:ac}|\gls{not:state})$ determines the probability of taking charging action \gls{not:ac} given network state \gls{not:state} which models the \gls{mc}'s behavior at a given time. We aim to find a policy $\gls{not:policy}[^*]$ that maximizes the expected $\gamma$-discounted network's lifetime.

We delineate a policy approximation using a deep neural network and a policy gradient algorithm to train the model in the following sections.

\subsection{Model Architecture} \label{sec:model_arc}

\begin{figure}[tb]
    \centering
    \includegraphics[width=0.8\linewidth]{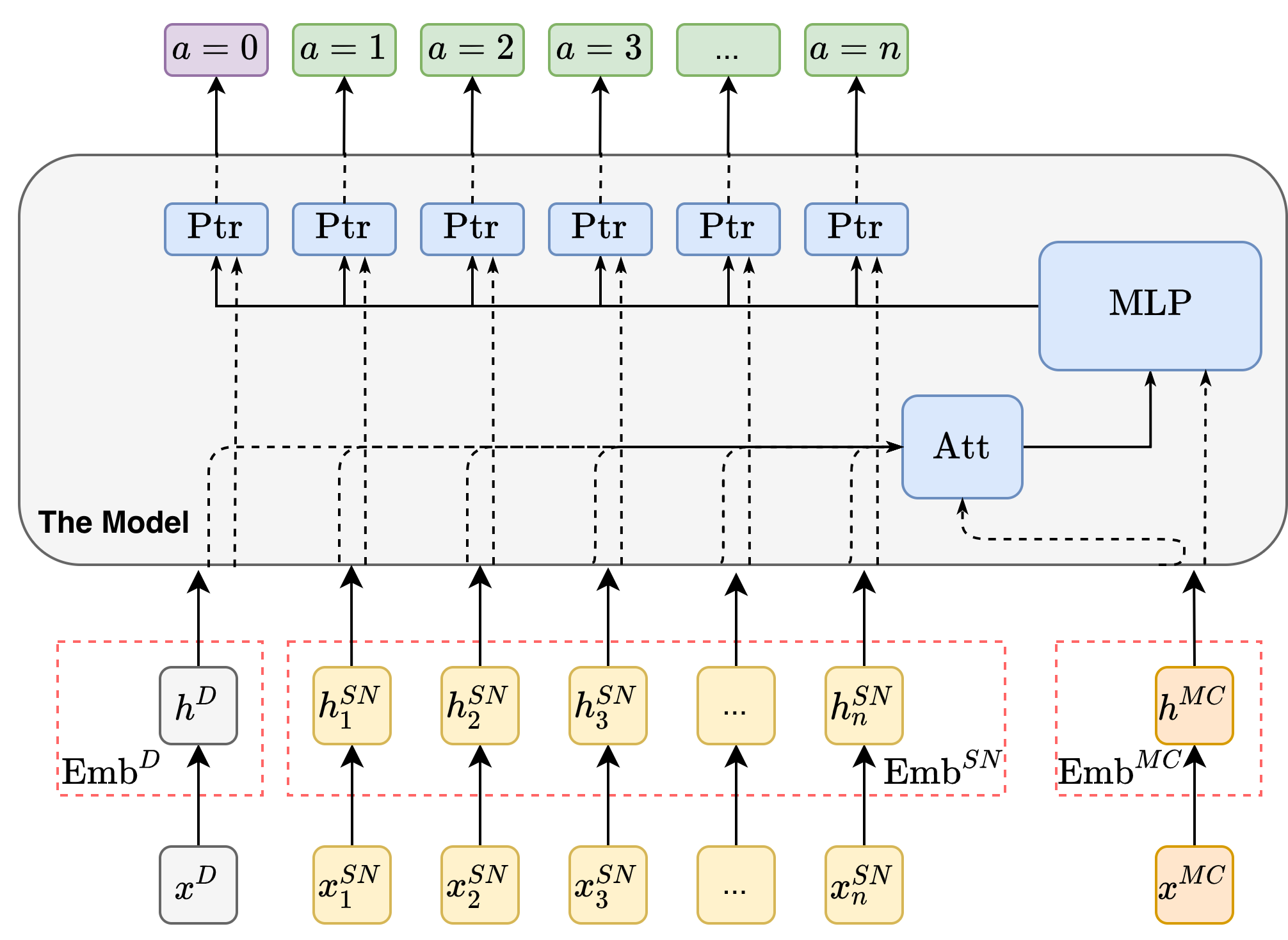}
    \caption{The model architecture of the actor.}
    \vspace{-5mm}
    \label{fig:model_arc}
\end{figure}

We parameterize the stochastic policy by $\gls{not:policy}[_\theta] (a|x)$, where $\theta$ is the parameters of a deep neural network aided by the attention and pointing mechanisms, similar to \citep{hottung2019neural}. Fig.~\ref{fig:model_arc} depicts the overall architecture.

The input to the model is the network state $\gls{not:state}_t$ at time $t$ which is composed of the states at time $t$ of the \gls{mc} $\gls{not:state:mc}_t$, the depot $\gls{not:state:depot}$, and the sensors $\bar{x}^{SN}_t$. The output of the model is the probabilities of taking action $a_t$ given the state $x_t$. To simplify the notation, in this section, we omit the timestamp $t$ when referring to the states $x_t$, $\gls{not:state:mc}_t$, and $\bar{x}^{SN}_t$.

We use three transformations $\Emb^{MC}$, $\Emb^D$, and $\Emb^{SN}$ to embed the input of the mobile charger, the depot, and the sensors, respectively, to a $d$-dimensional latent space. Note that we use the same transformation $\Emb^{SN}$ for all sensors and the embedding vector of each sensor is computed separately and identically. Let $h^{MC}, h^D, h^{SN}_i \in \R^d$ be the embedded input corresponding to $x^{MC}, x^D, x^{SN}_i$. For convenience's sake, we denote $\bar{h}^C = \{h^D, h^{SN}_1, ...h^{SN}_n\}$ as a matrix of embedded input of charging destinations and then refer to $\bar{h}^C_i$ as the embedding of charging destination $i$. An attention layer is used to extract alignment vector $\bar{a}$, which specifies how much `attention' the \gls{mc} might have for each charging destination given their current status. Precisely, the alignment vector is calculated by the following formula:
\begin{equation}
    \bar{a} = \softmax(u^H_0, u^H_1, ..., u^H_n),
\end{equation}
where:
\begin{equation}
    u^H_i = z^A \tanh{(W^A [\bar{h}_i^C; h^{MC}])}.
\end{equation}
Here, $[;]$ denotes the concatenation of two vectors. The context vector $c$ is provided by:
\begin{equation}
    c = \sum_{i=0}^n \bar{a}_i \bar{h}_i^C.
\end{equation}

The context vector is later concatenated with the embedded input of the \gls{mc} to be the input of a multilayer perceptron ($\mathrm{MLP}$) with one hidden layer that outputs a vector $q \in \R^d$. 
\begin{equation}
    q = \mathrm{MLP}_{W^B} ([c;h^{MC}]).
\end{equation}

The distribution of the policy over all actions for the state $x$ is then given by:
\begin{equation}
    \label{eq:ptr}
    \gls{not:policy}[_\theta] (a = i | x) = \softmax(u_0, u_1, ..., u_n),
\end{equation}
where
\begin{equation}
    u_i = z^C \tanh{(\bar{h}^C_i + q)},
\end{equation}
and $\theta = \{z^A, W^A, W^B, z^C\}$ are trainable parameters.

The pointing mechanism~\cite{vinyals2015pointer} in the Eq.~\eqref{eq:ptr} is a reduction of attention mechanism that leverages the alignment vector to determine the probabilities of selecting each member in the input. Both only require learning a tuple of parameters to compute utilities of charging actions ($(z^A, W^A)$ for attention and $z^C$ for pointing). It thus imposes the permutation invariant property for the set of the input's members, i.e., the output probabilities of the policy depend solely on the features of the sensors but not the order of the sensors in the input state. Additionally, it allows the MC to operate on a dynamic network where the number of sensors might change due to node failures or node deployments. It differs from the prior DRL work for the on-demand approach~\cite{cao2021deep}, which requires modifying the architecture and retraining the model when the number of sensors is changed. Moreover, instead of using a Gated Recurrent Unit (GRU) as in the pointer network~\cite{vinyals2015pointer}, we use a fully connected MLP, similar to~\cite{hottung2019neural}, to avoid the dependence of the \gls{mc}'s decision on the previous state. It enables the trained MC to be deployed on the fly into an existing WSN.

\begin{algorithm}
\caption{Policy gradient algorithm}
\label{algo:training}
\begin{algorithmic}[1]
\REQUIRE A set of network instances $\mc D$, discount factor $\gamma$, GAE hyperparameter $\lambda$, regularization hyperparameter $\beta$.
\ENSURE A trained policy $\pi_\theta$.
\STATE initialize the actor network with random weight $\theta$.
\STATE initialize the critic network with random weight $\psi$.

\FOR{$\mathrm{epoch} = 1, 2, ...$}
    \FOR{$n = 1, ..., N$}
        \STATE initialize the environment on the instance $D_n \in \mc D$.
        \STATE generate an episode following $\pi_\theta$: 
        \\ $\quad x_0, a_0, x_1, a_1,..., x_{T-1}, a_{T-1}, x_{T}$
        \STATE $G \gets 0, A_t^{\mathrm{GAE}} \gets 0$
        \STATE $d\theta \gets 0, d\psi \gets 0$
        \FOR{$t = T-1, T-2, ..., 0$}
            \STATE $G \gets \gamma G + R_{t}$
            \STATE $\delta_t \gets R_t + \gamma V_\psi(x_{t+1} - V_\psi(x_t)$
            \STATE $A_t^{\mathrm{GAE}} \gets \gamma\lambda A_t^{\mathrm{GAE}} + \delta_t$
            \STATE $\diff\theta \gets d\theta - \nabla_\theta \log(\pi_\theta(a_t|x_t) A_t^{\mathrm{GAE}} - \beta \nabla_\theta H(\pi_\theta(\cdot|x_t))$
            \STATE $d\psi \gets d\psi + \nabla_\psi \frac{1}{2}  \left\lVert G - V_\psi(x_t)\right\rVert _2^2$
        \ENDFOR
        \STATE $\theta \gets \mathrm{Adam}(\theta, d\theta)$
        \STATE $\psi \gets \mathrm{Adam}(\psi, d\psi)$
    \ENDFOR
\ENDFOR
\end{algorithmic}
\end{algorithm}

\subsection{Policy Optimization} \label{sec:training}

We train the agent using a well-known policy gradient method in reinforcement learning. Our objective is to maximize the expected network's lifetime:
\begin{equation}
    J(\theta) = \mathbb{E}_{\gls{not:traj} \sim p_\theta(\gls{not:traj})} \Bigg[\sum^{T}_{t=0} \gls{not:gamma}^t R(x_t, a_t) \Bigg]
    \label{eq:obj},
\end{equation}
where $p_\theta(\gls{not:traj})$ is the distribution of the Markov chain induced by the policy $\gls{not:policy}[_\theta]$ which generates the trajectories \gls{not:traj}. Applying the REINFORCE~\citep{williams1992simple}, the gradient of Eq.~\eqref{eq:obj} can be computed as:
\begin{equation}
    \nabla J(\theta) = \mathbb{E}_{\gls{not:traj} \sim p_\theta(\gls{not:traj})} \Bigg[\sum^{T}_{t=0} \nabla_\theta \log{(\pi_\theta (a_t | x_t))} A_t \Bigg],
\end{equation}
where $A_t = \sum_{l=t}^{T} \gamma^{l-t} R(x_{t}, a_{t})  - V_\psi(x_t)$ is the advantage function of taking action $a_t$ given state $x_t$. This vanilla policy gradient update has no bias but high variance, which leads to an unstable learning process. To overcome this issue, we use the Generalized Advantage Estimation (GAE)~\citep{schulman2015high} to reduce the variance caused by the original advantage function at the cost of introducing bias. Furthermore, due to the non-convex objective function, the policy gradient usually suffers from local convergence. We hence use the entropy regularization as suggested in~\citep{mnih2016asynchronous} to encourage the exploration. The following formula computes the final policy gradient: 
\begin{equation}\label{eq:grad}
\begin{split}
    \nabla J(\theta) = \mathbb{E}_{\gls{not:traj} \sim p_\theta(\gls{not:traj})} \Bigg[& \sum^{T}_{t=0} \nabla_\theta \log{(\pi_\theta (a_t | x_t))} \hat{A}^{\mathrm{GAE}(\lambda)}_t \\ & + \beta \nabla_\theta H(\pi_\theta(\cdot|x_t))  \Bigg],
\end{split}
\end{equation}
where $H$ is the entropy function, $\beta$ is a hyperparameter controlling the strength of the regularization, and $\hat{A}^{\mathrm{GAE}(\lambda)}_t$ is the GAE function, which is estimated by:
\begin{equation}
    \hat{A}^{\mathrm{GAE}(\lambda)}_t = \sum_{l=t}^{T} (\gamma\lambda)^l (R_{l} + \gamma V_\psi(x_{l+1}) - V_\psi(x_{l})),
\end{equation}
where $\lambda$ is a hyperparameter controlling the trade-off between variance and bias in the advantage function. The function $V_\psi(x)$ is the estimated value function at the state $x$. 

During training, we maintain two networks, one for the policy $\gls{not:policy}[_\theta](a|x)$, and another for value function $V_\psi(x)$ with the trainable parameter $\theta$ and $\psi$, respectively. We use a simple three-layer fully connected MLP with a ReLU activation in between for the value function's network. The policy is updated using gradients computed in Eq.~\eqref{eq:grad}. Meanwhile, we use Mean Square Error (MSE) to compute the loss function for the value function. Both networks are trained with \textit{Adam} optimizer \citep{kingma2014adam}. The pseudocode of our training process is described in Algorithm~\ref{algo:training}. 

\begin{figure*}[!hbt]
    \raggedleft
    \renewcommand{\thesubfigure}{\alph{row}\arabic{subfigure}}%
    \setcounter{row}{1}%
    \begin{subfigure}[t]{0.28\textwidth}
        \raggedleft
        \includegraphics[width=\textwidth]{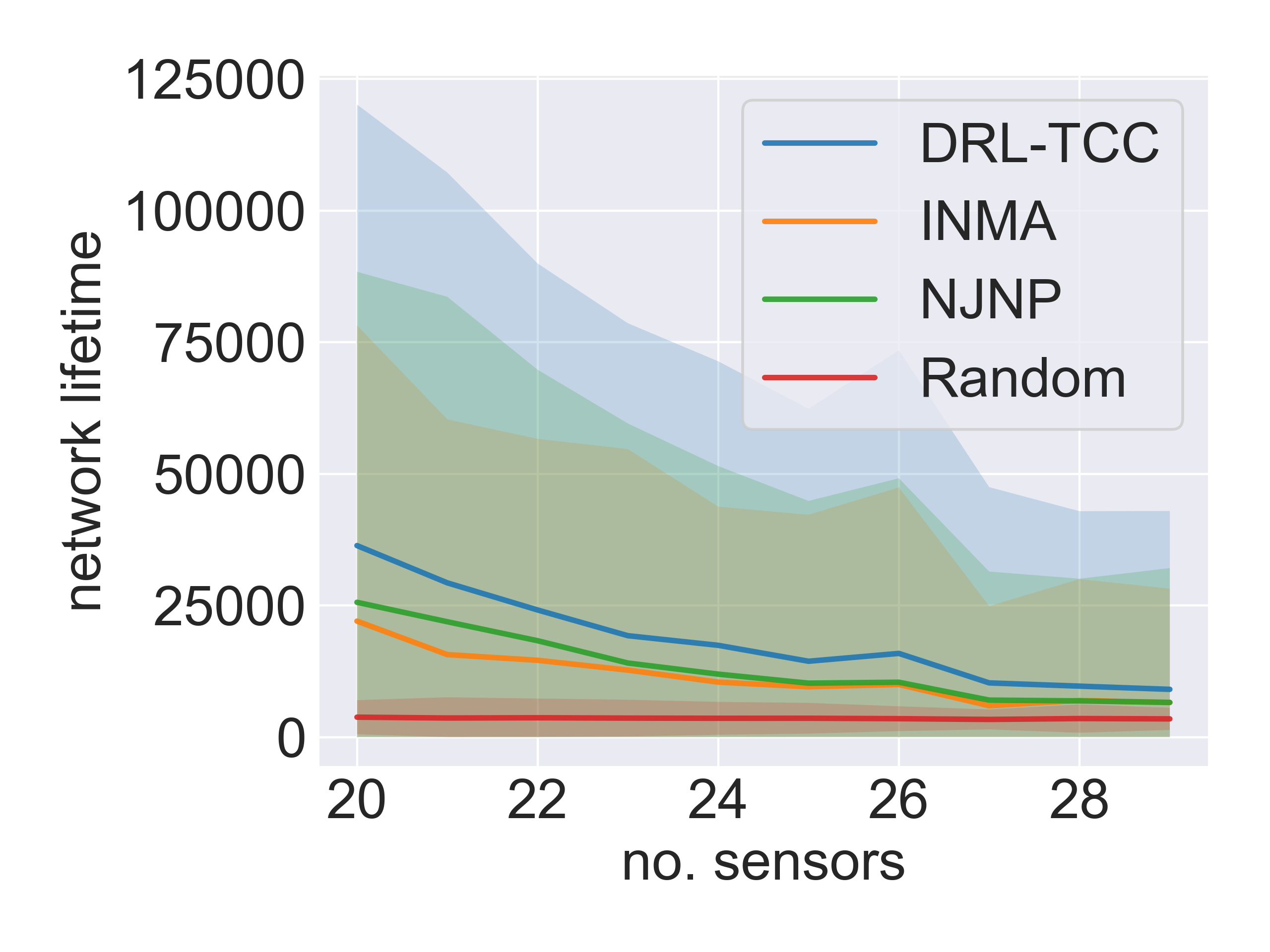}
        \vspace{-7mm}
        \caption{}
        \label{fig:lifetime:1a}
    \end{subfigure}
    \hfill
    \begin{subfigure}[t]{0.28\textwidth}
        \raggedleft
        \includegraphics[width=\textwidth]{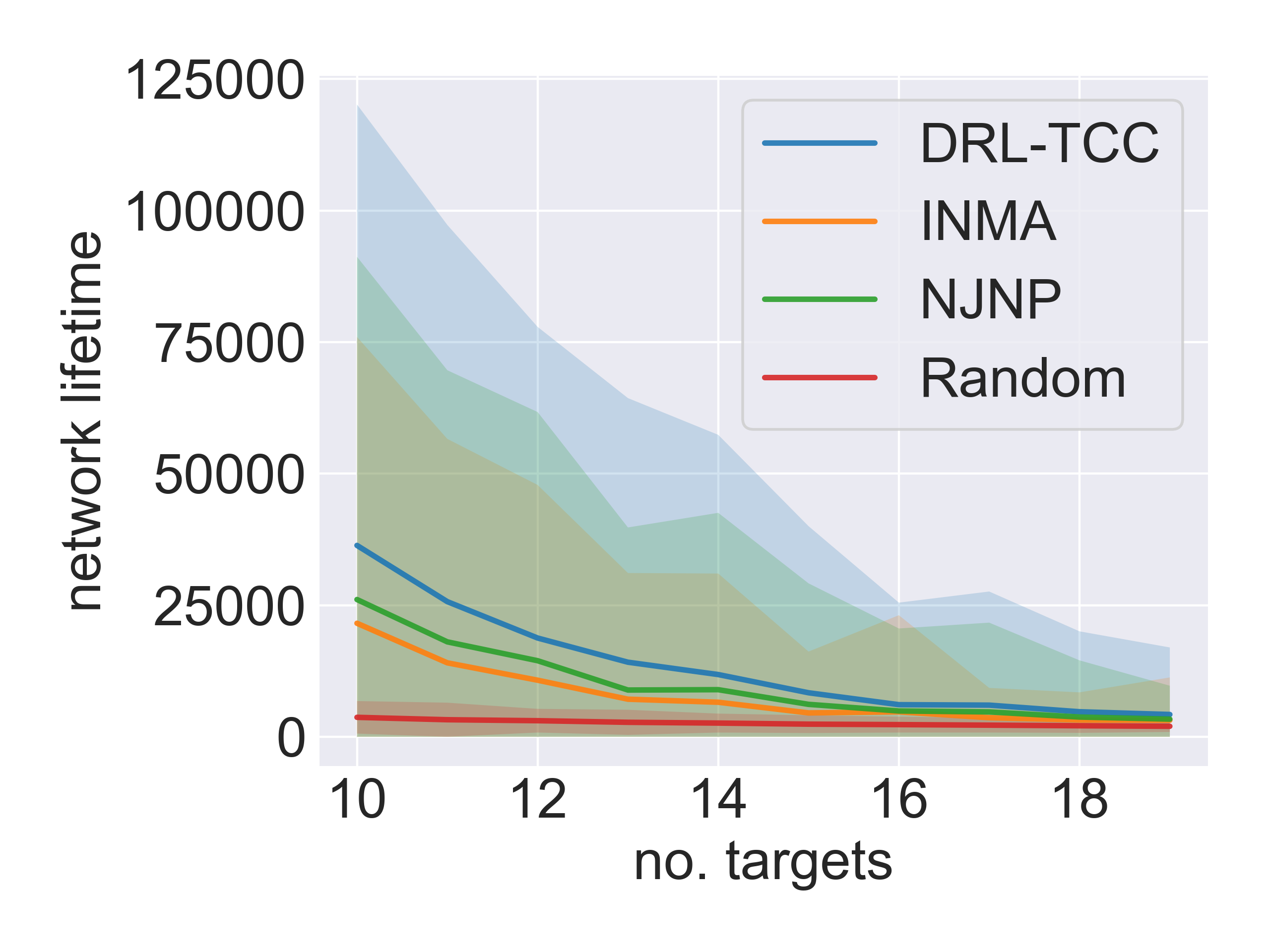}
        \vspace{-7mm}
        \caption{}
        \label{fig:lifetime:1b}
    \end{subfigure}
    \hfill
    \begin{subfigure}[t]{0.28\textwidth}
        \raggedleft
        \includegraphics[width=\textwidth]{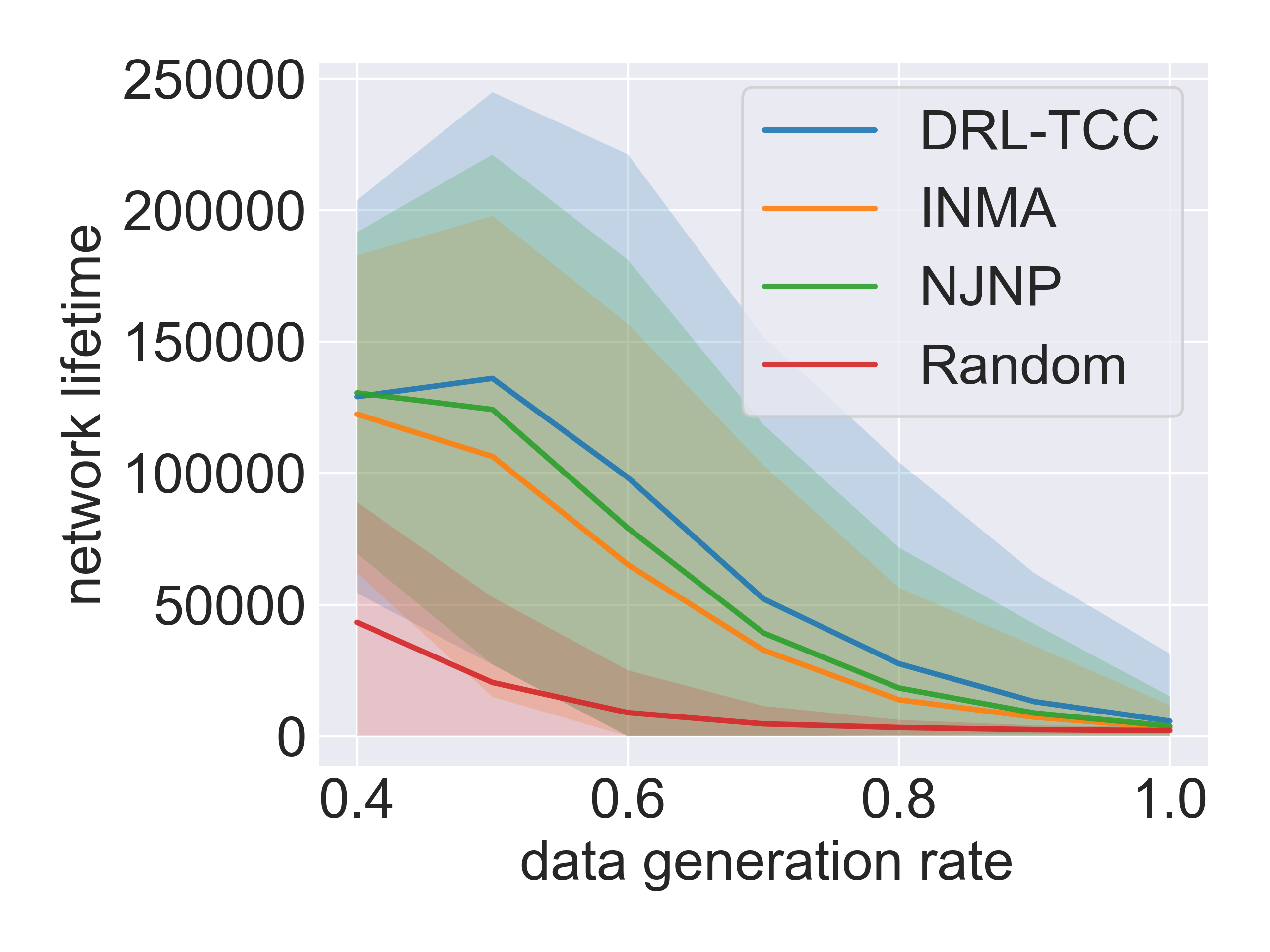}
        \vspace{-7mm}
        \caption{}
        \label{fig:lifetime:1c}
    \end{subfigure} 
    
    \setcounter{row}{2}%
    \setcounter{subfigure}{0}
    
    \begin{subfigure}[t]{0.28\textwidth}
        \raggedleft
        \includegraphics[width=0.93\textwidth]{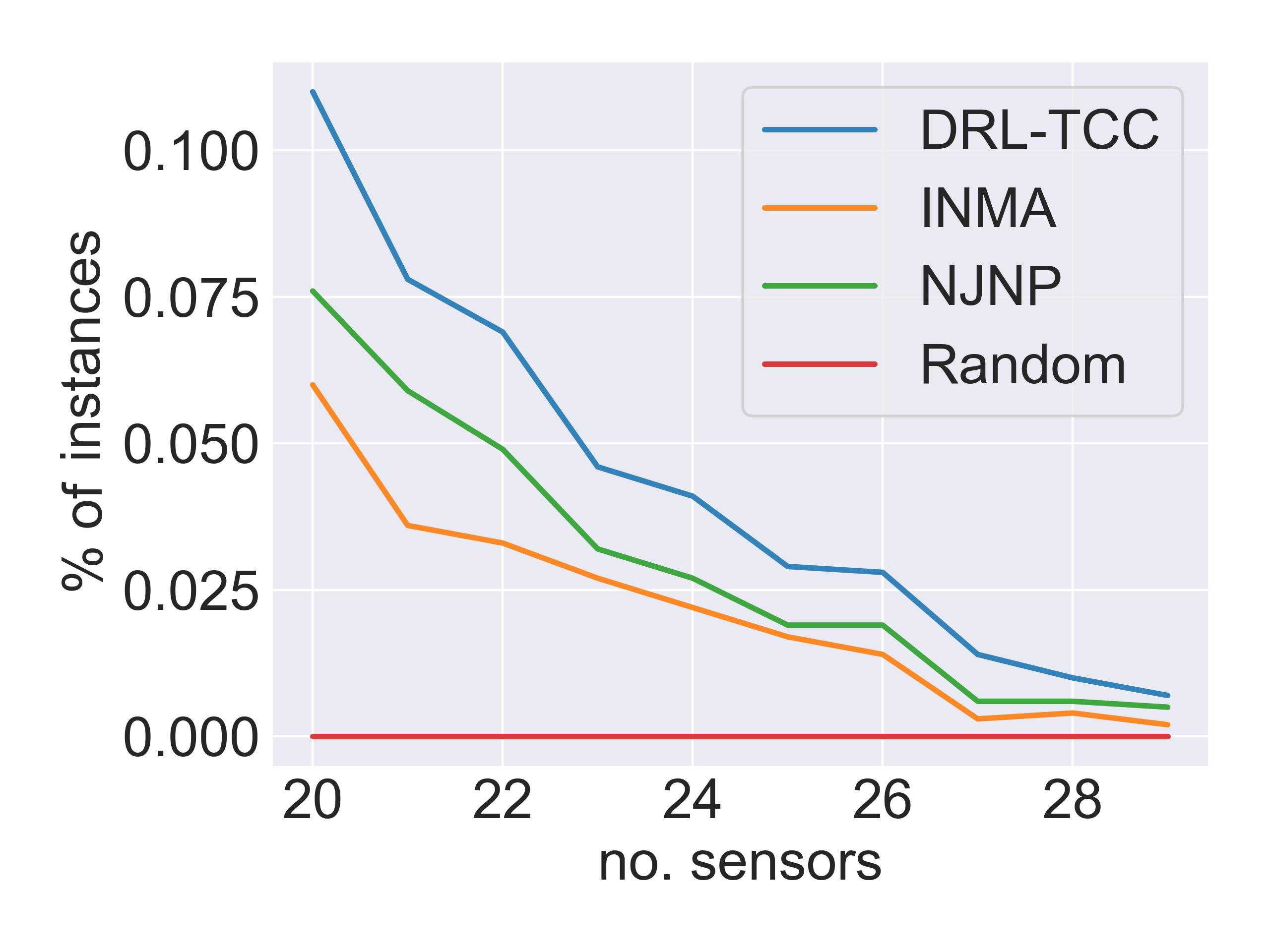}
        \vspace{-2mm}
        \caption{}
        \label{fig:lifetime:2a}
    \end{subfigure}
    \hfill
    \begin{subfigure}[t]{0.28\textwidth}
        \raggedleft
        \includegraphics[width=0.93\textwidth]{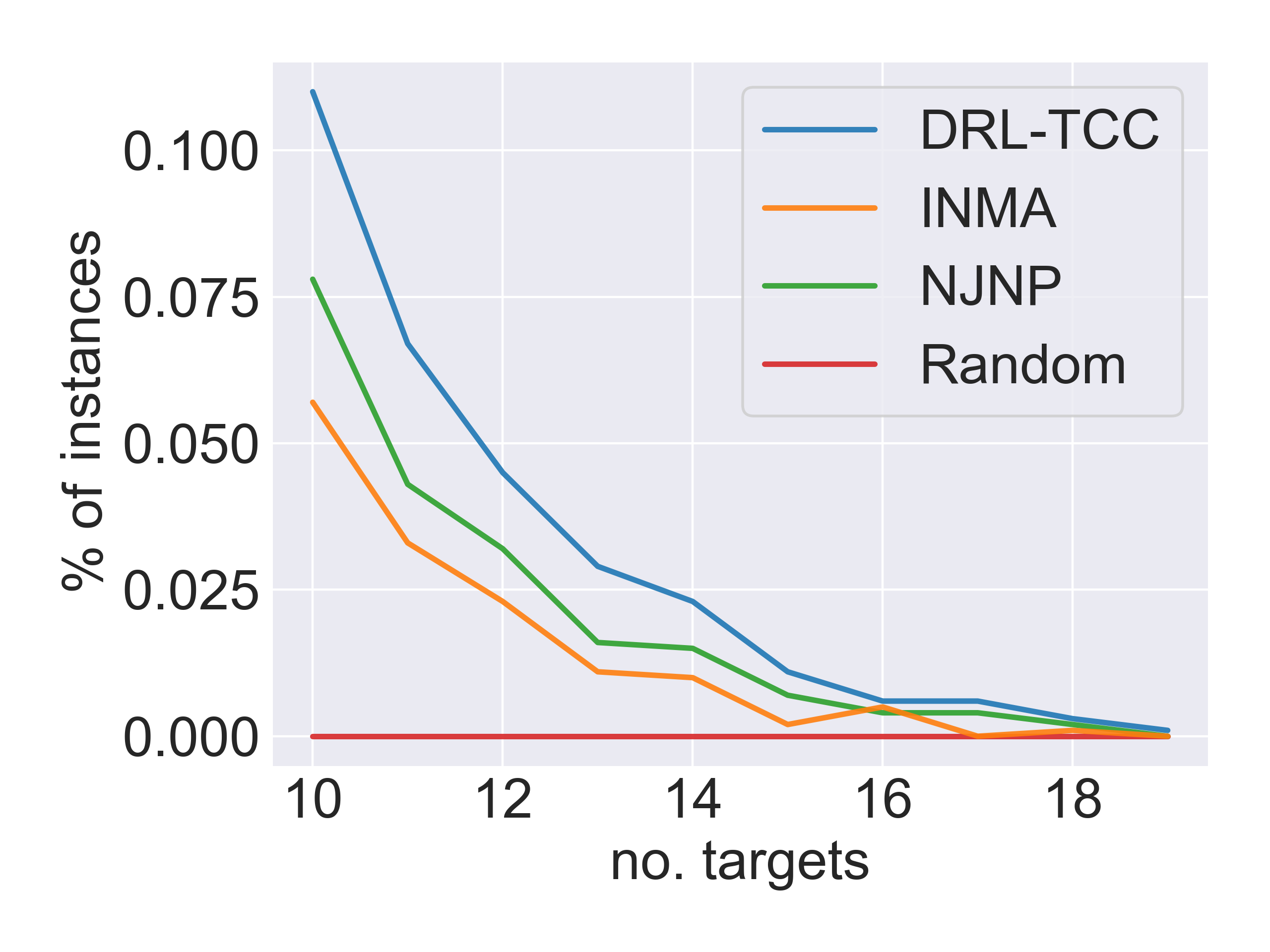}
        \vspace{-2mm}
        \caption{}
        \label{fig:lifetime:2b}
    \end{subfigure}
    \hfill
    \begin{subfigure}[t]{0.28\textwidth}
        \raggedleft
        \includegraphics[width=0.93\textwidth]{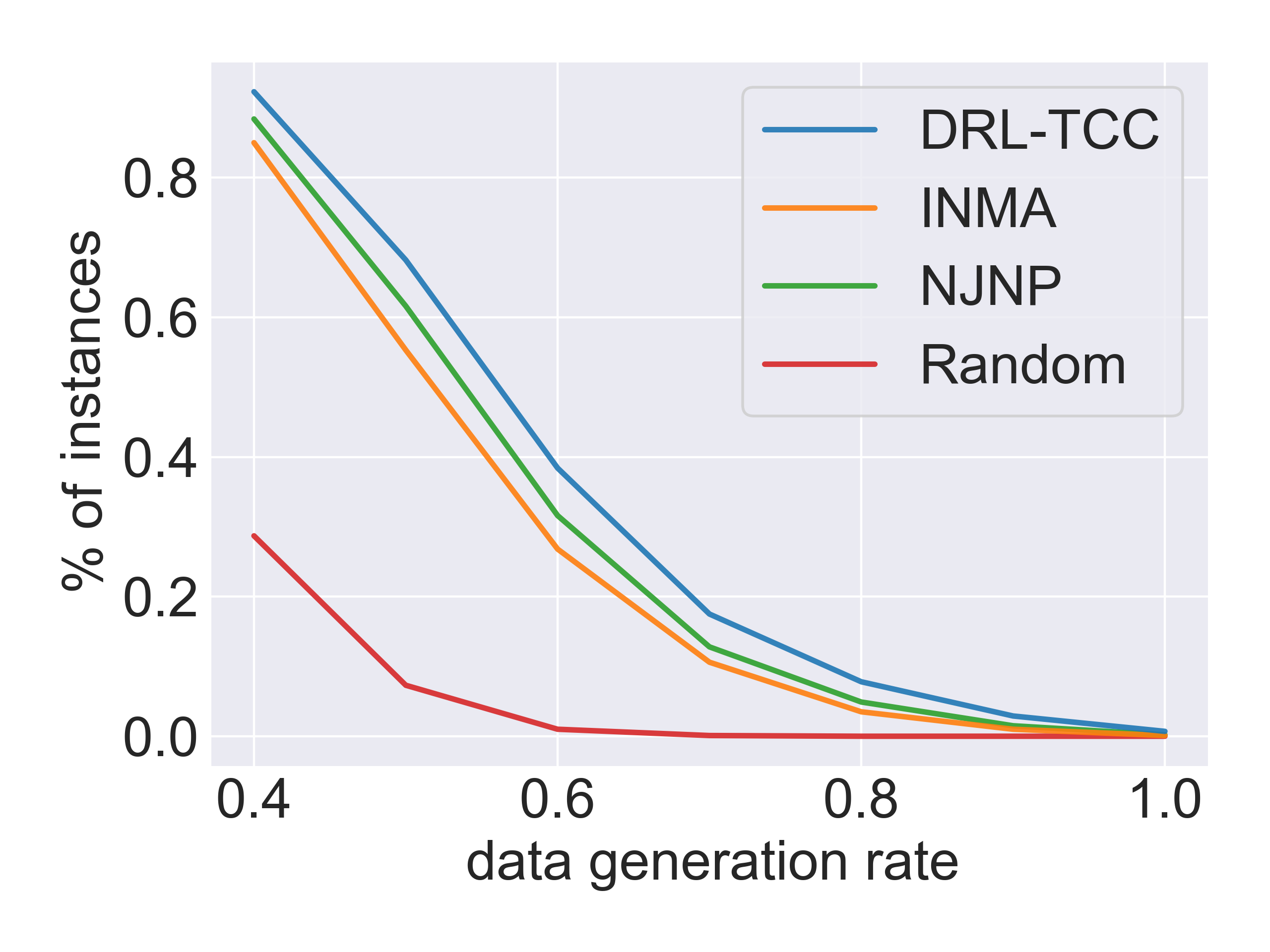}
        \vspace{-2mm}
        \caption{}
        \label{fig:lifetime:2c}
    \end{subfigure}
    \caption{Evaluation of the network lifetime of competing algorithms when varying the number of sensors, number of targets, or package generation probability.}
    \vspace{-5mm}
    \label{fig:lifetime}
\end{figure*}

\section{Empirical Results}\label{sec:expt}
In this section, we conduct simulations to demonstrate the merits of our method compared to existing on-demand charging algorithms. We also discuss the self-organizing capability of the wireless sensor networks powered by our mobile charger. The model and the environment simulation are implemented using the PyTorch and Gym framework. All experiments are performed in a computer with an Intel(R) Core(TM) i7-6800K CPU @ 3.40GHz, 16 GB RAM, and GeForce GTX TITAN X GPU running on Ubuntu Linux 18.04.

\subsection{Simulation Settings}

\textbf{Network structure. } 
We assume that the sensor network is randomly deployed in a square area of interest $200 \times 200$. The sink is supposed to be located in the center $(100, 100)$ of the field, and the depot is in the bottom-left corner $(0,0)$. Initially, all sensors are equipped with a full and rechargeable battery with the capacity of $\gls{not:c:sn} = 10J$. Meanwhile, the \gls{mc}, which has a high-capacity battery ($\gls{not:c:mc} = 500J$), is initially located at the depot with an initial energy of $50J$. The reason for the low initial power setting is to enforce the exhausted state of the \gls{mc} in some first states, which encourages the \gls{mc} to learn when to go back depot for recharging in some first episodes. That, in turn, accelerates the learning process. We set the other parameters as in Table~\ref{tab:network_parameters}, similar to~\citep{he2013demand}.

\textbf{Implementation. }
To train a DRL model, we generate $10000$ network instances with $20$ sensors and $10$ targets. The positions of sensors and targets are drawn in a square area according to the uniform distribution. We assume that all instances are guaranteed to be covered and connected initially. The probability of a sensor sending a packet of sensing data in a unit time period $\kappa$ (data generation rate) is set at $0.8$. For the training parameters, we set the reward discount $\gls{not:gamma}$, the entropy regularization $\beta$, and GAE $\lambda$ to $0.95$, $0.02$, and $0.9$, respectively. We then train one DRL model in these environments and fix the agent throughout the evaluations.

\begin{table}[tb!]
  \centering
  \caption{Configuration of the simulations.}
  \label{tab:network_parameters}
  \footnotesize
  \begin{tabular}{llll}
    \toprule
    Parameter & Value & Unit & Comment \\
    \midrule
    \gls{not:num:sn} & $20 \sim 30$ & $-$ & \glsdesc*{not:num:sn} \\
    \gls{not:num:tg} & $10 \sim 20$ & $-$ & \glsdesc*{not:num:tg} \\
    \gls{not:c:mc} & $500$ &  $J$ & \glsdesc*{not:c:mc} \\
    \gls{not:e:move} & $0.04$ & $J/m$ & \glsdesc*{not:e:move}\\
    \gls{not:v} & $5$ & $m/s$ & \glsdesc*{not:v}\\
    \gls{not:c:sn} & $10$ & $J$ & \glsdesc*{not:c:sn}\\
    \gls{not:r:s} & $40$ & $m$ & \glsdesc*{not:r:s}\\
    \gls{not:r:c} & $80$ & $m$ & \glsdesc*{not:r:c}\\
    \gls{not:mu} & $0.04$ & $J/s$ & \glsdesc*{not:mu}\\
    \bottomrule
  \end{tabular}
  \vspace{-5mm}
\end{table} 

\textbf{Baselines. }
We mainly compare our method, namely DRL-TCC, against the existing on-demand charging algorithms:

\begin{itemize}
    \item Random: The agent chooses the next charging destination at random. We add a simple estimation that helps the agent go back to the depot to recharge itself before being exhausted.
    \item NJNP \citep{he2013demand}: A heuristic algorithm with a simple but very efficient discipline that chooses the spatially closest requesting node as the next charging destination. 
    \item INMA \citep{zhu2018adaptive}: A modified algorithm of NJNP that selects nodes that make the least number of other requesting nodes enduring energy deficiency as the charging candidates. For high charging efficiency, the node with the shortest time to finish the charging will be selected as the next charging node if the candidate set has more than one node.
\end{itemize}

\subsection{Performance Evaluation}
\begin{figure*}[!hbt]
    \centering
    \renewcommand{\thesubfigure}{\alph{row}\arabic{subfigure}}%
    
    \setcounter{row}{1}%
    \begin{subfigure}[t]{0.28\textwidth}
        \centering
        \includegraphics[width=\textwidth]{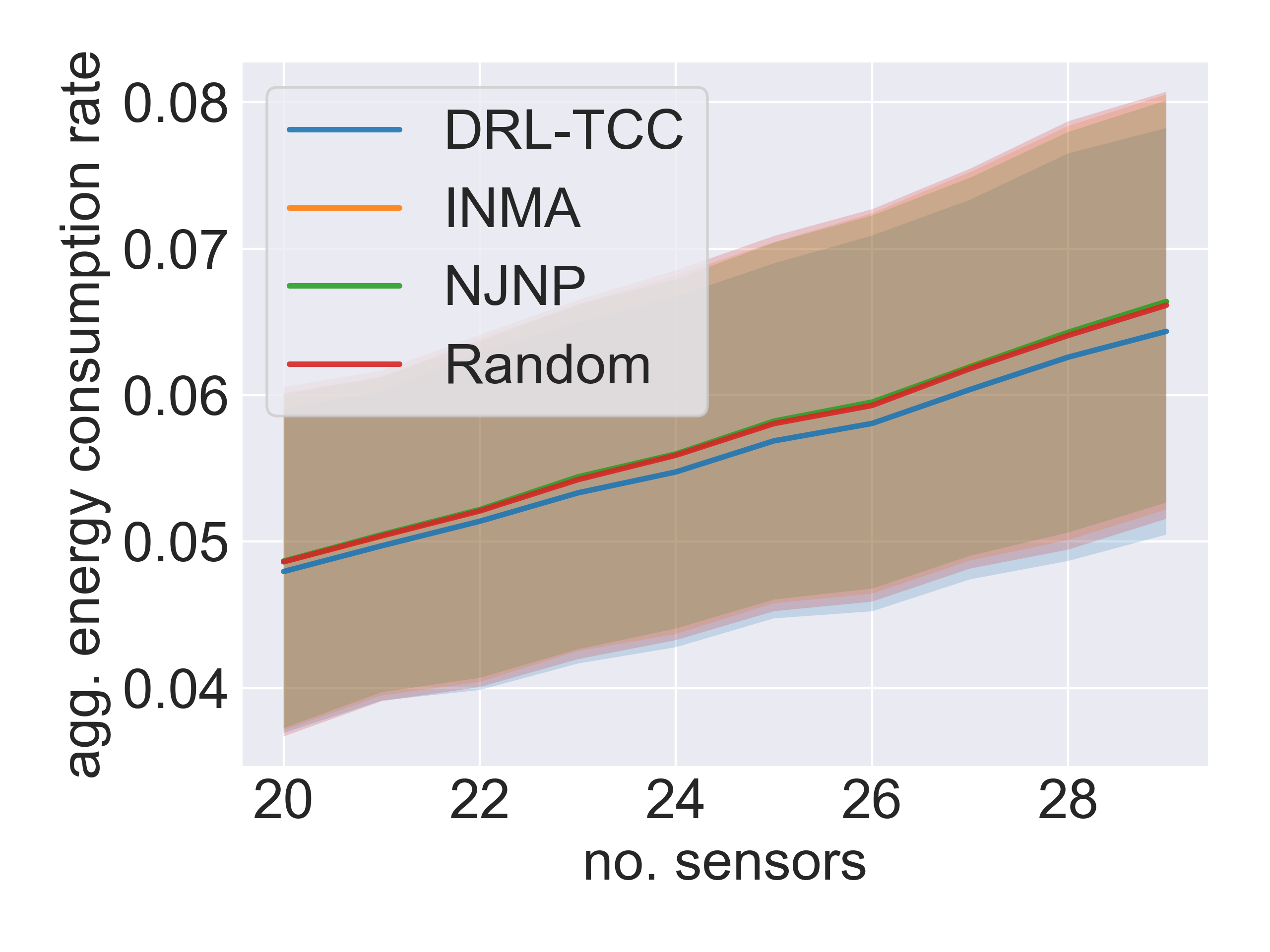}
        \vspace{-7mm}
        \caption{}
    \end{subfigure}
    \hfill
    \begin{subfigure}[t]{0.28\textwidth}
        \centering
        \includegraphics[width=\textwidth]{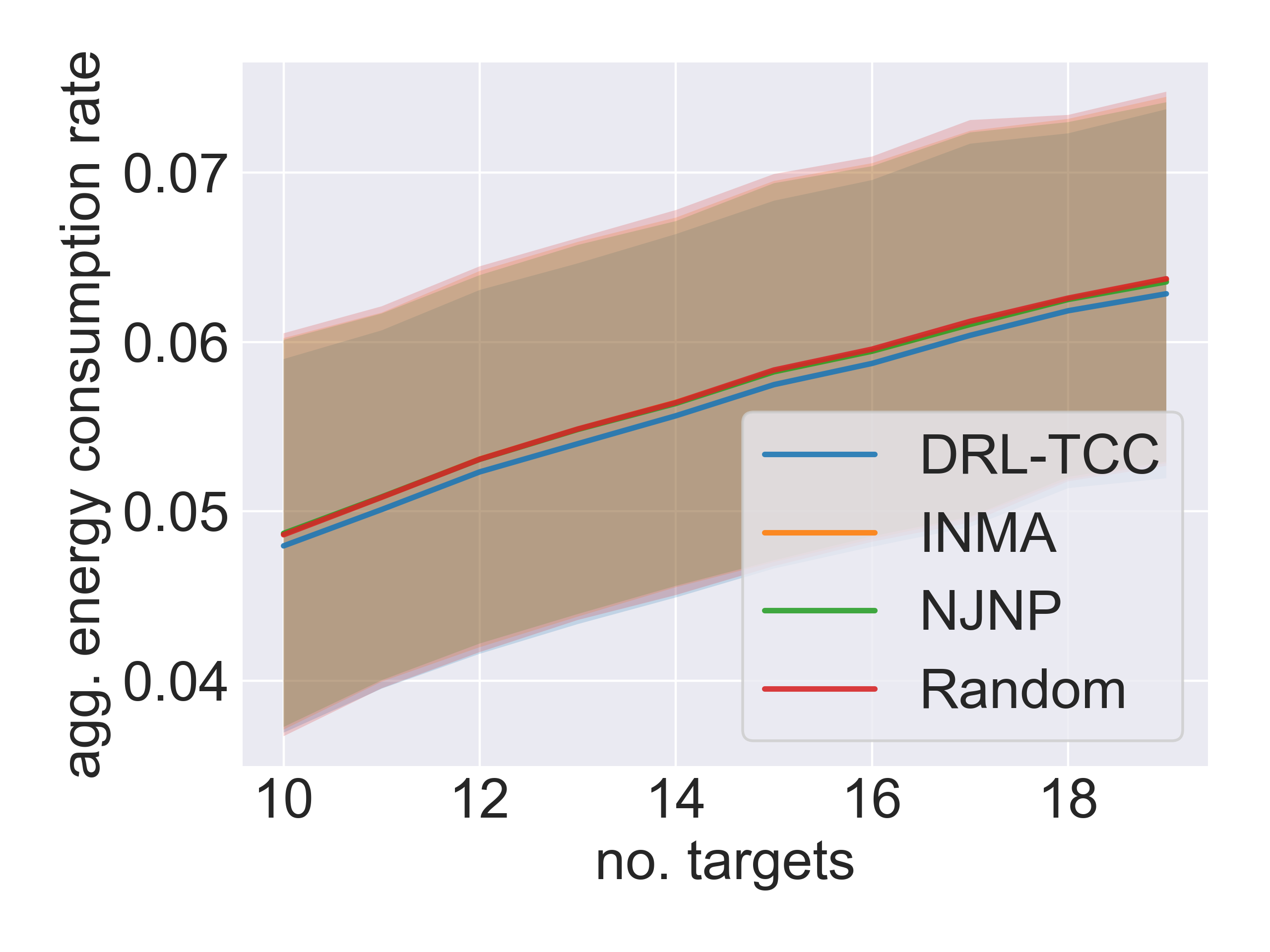}
        \vspace{-7mm}
        \caption{}
    \end{subfigure}
    \hfill
    \begin{subfigure}[t]{0.28\textwidth}
        \centering
        \includegraphics[width=\textwidth]{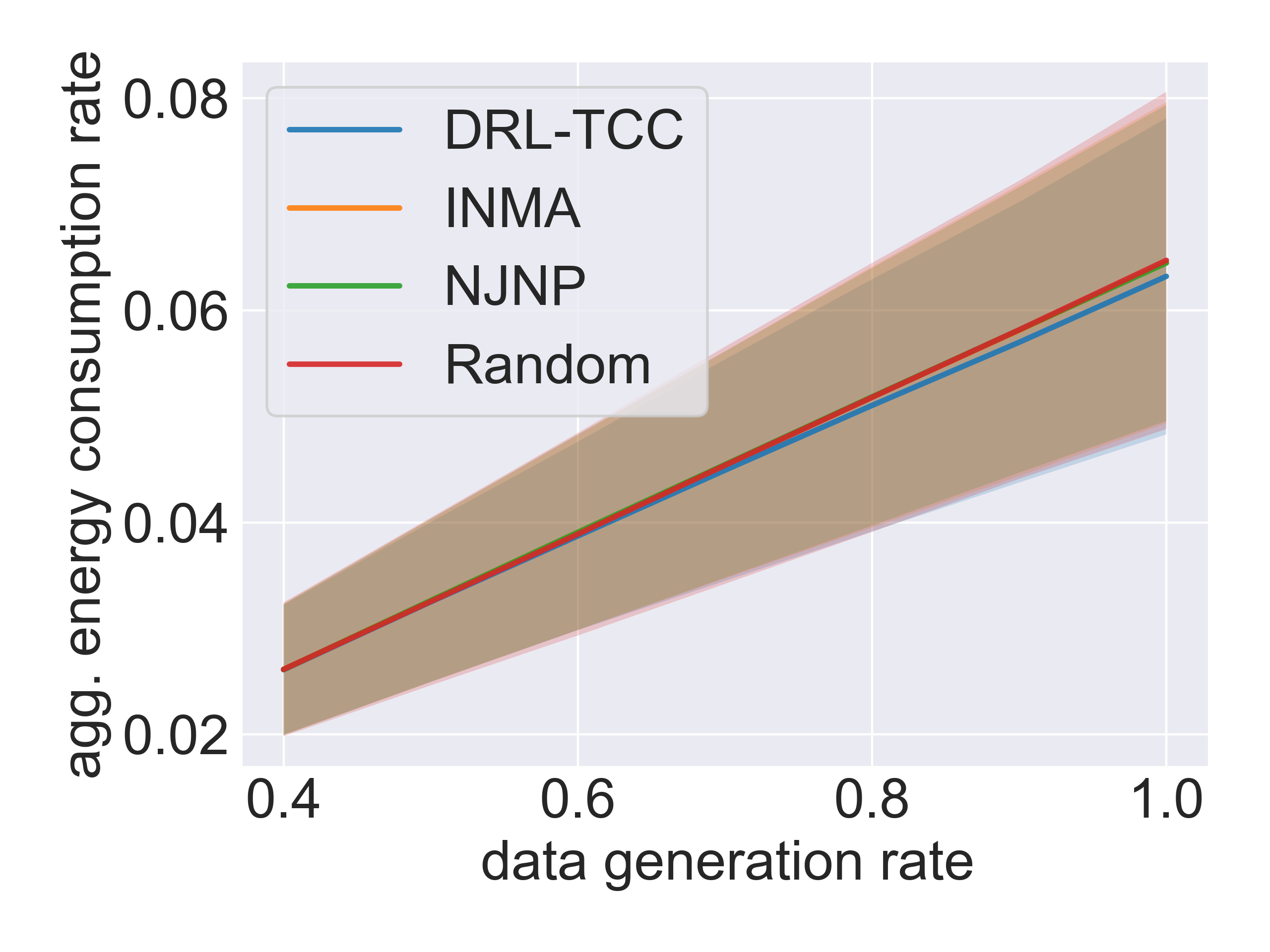}
        \vspace{-7mm}
        \caption{}
    \end{subfigure} 
    \setcounter{row}{2}%
    \setcounter{subfigure}{0}
    \begin{subfigure}[t]{0.28\textwidth}
        \centering
        \includegraphics[width=\textwidth]{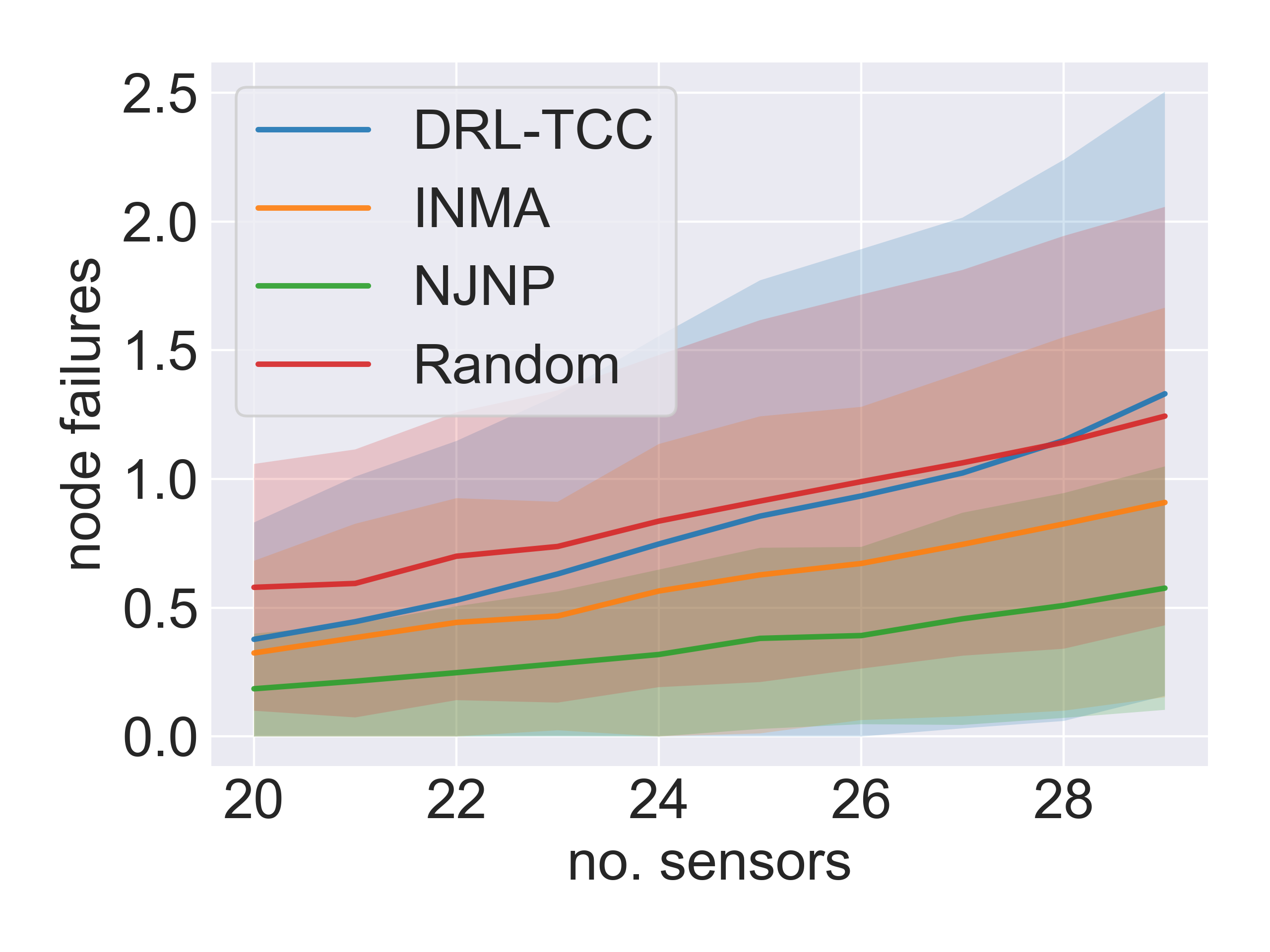}
        \vspace{-7mm}
        \caption{}
    \end{subfigure}
    \hfill
    \begin{subfigure}[t]{0.28\textwidth}
        \centering
        \includegraphics[width=\textwidth]{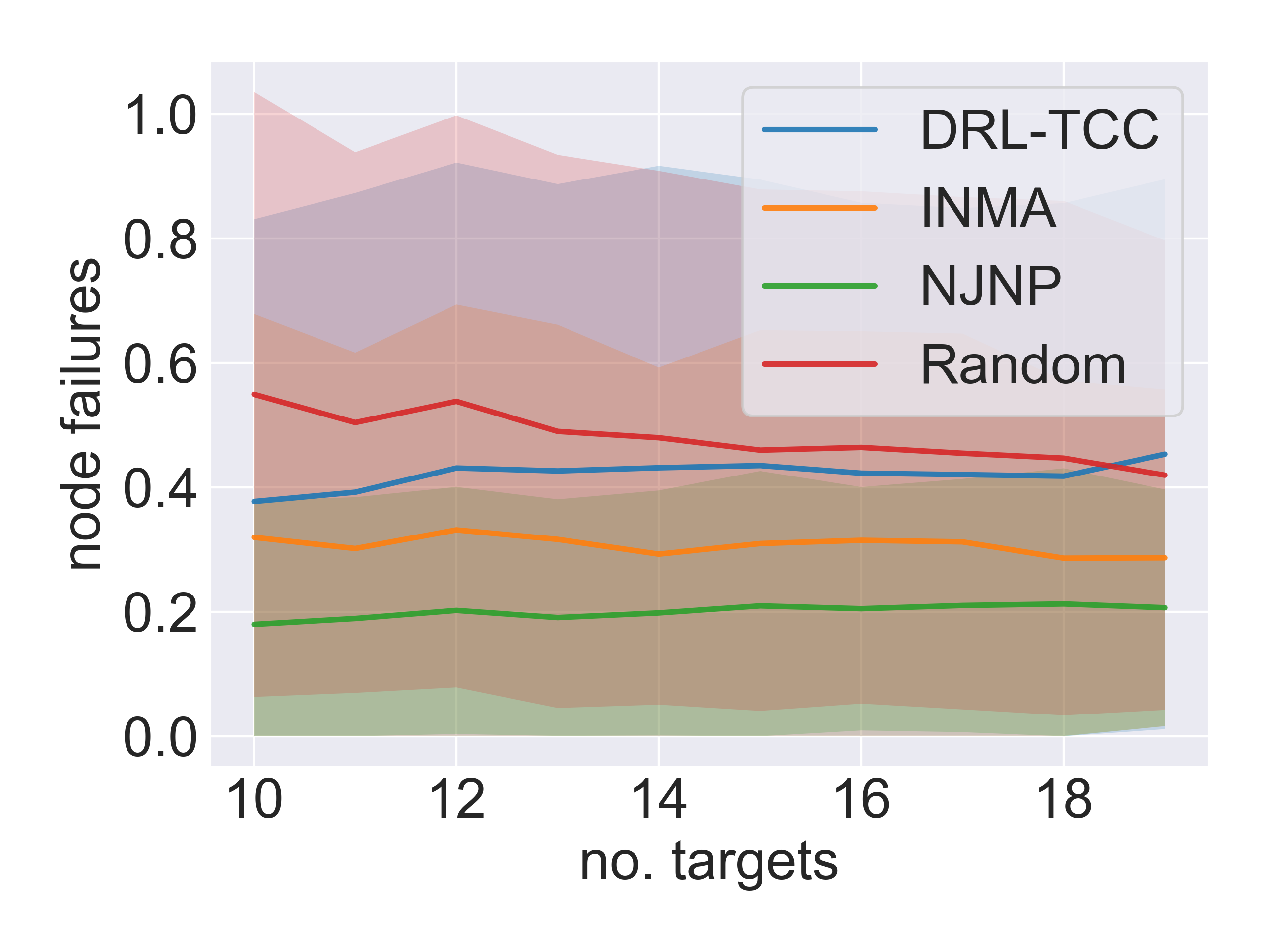}
        \vspace{-7mm}
        \caption{}
    \end{subfigure}
    \hfill
    \begin{subfigure}[t]{0.28\textwidth}
        \centering
        \includegraphics[width=\textwidth]{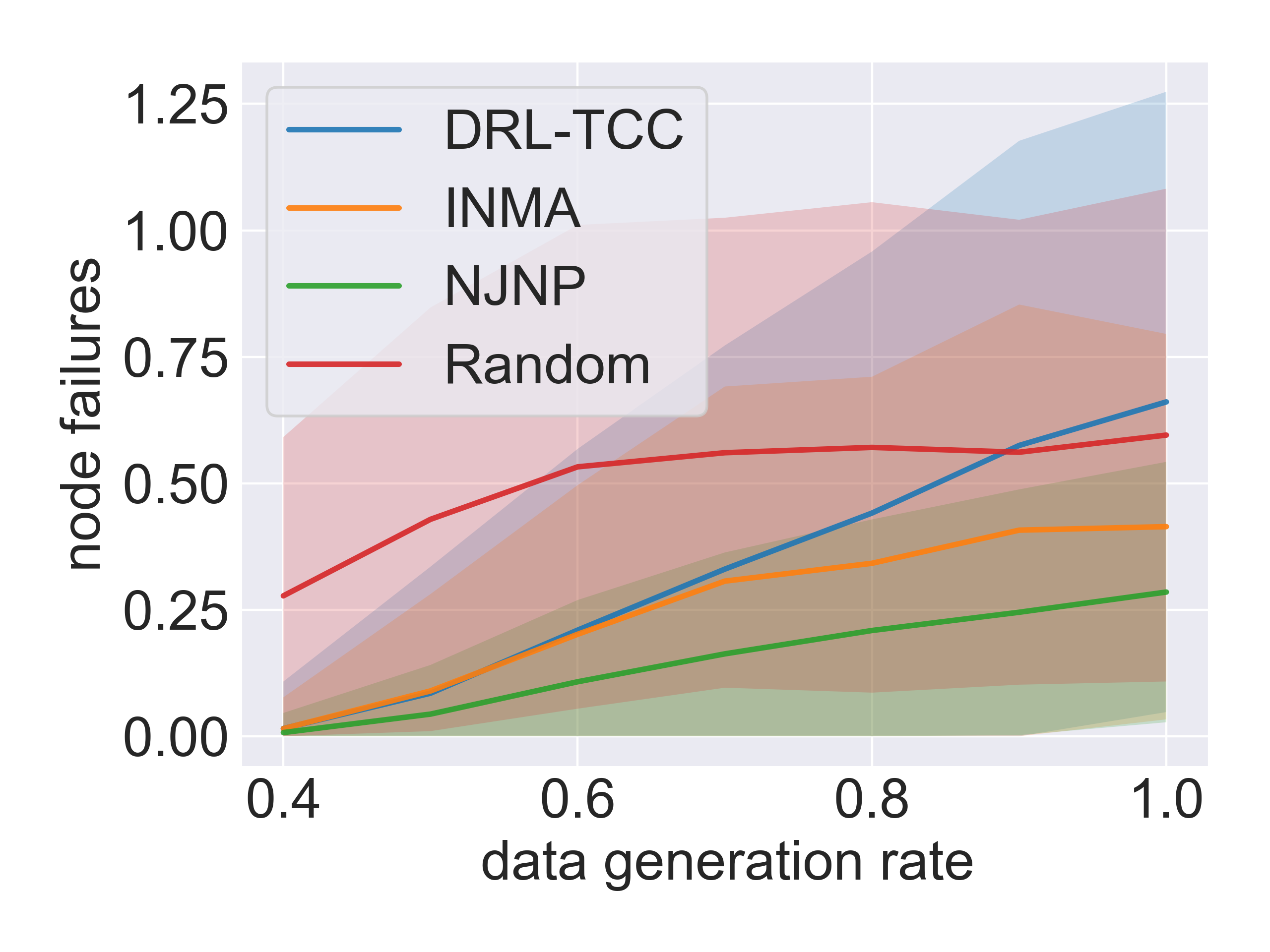}
        \vspace{-7mm}
        \caption{}
        \label{fig:node_failures:prob}
    \end{subfigure} 
    \caption{The comparison of the aggregated energy consumption rate and the number of node failures when increasing the number of sensors, number of targets, or data generation rate.}
    \vspace{-5mm}
    \label{fig:self-org}
\end{figure*}

We assess the ability to prolong the network's lifetime of the evaluated algorithms on various topologies. To understand the impact of the number of sensors ($n$), the number of targets ($m$), and the data generation rate ($\kappa$) on the performance of the competing algorithms, we vary one factor while leaving the other two as the default values ($n = 20$, $m = 10$, $\kappa = 0.8$). For each configuration, we generate $1000$ network instances where the sensors' and targets' positions are drawn from the same uniform distribution. We then simulate the WRSN operations and report the average cumulative lifetime over all instances. 

To prevent endless simulation, for each instance, we limit the maximum charging actions of the mobile charger to $2000$. We then report the number of instances with respect to which the system still sustains the full coverage and connectivity after having done the charging limit. The results are shown in Fig.~\ref{fig:lifetime}. 

\textbf{Impact of $n$, $m$.  } We vary the number of sensors from 20 to 30 (Figure~\ref{fig:lifetime:1a} and~\ref{fig:lifetime:1b}), and the number of targets from 10 to 20 (Figure \ref{fig:lifetime:2a} and~\ref{fig:lifetime:2b}). Generally, the network lifetime decreases when increasing the number of sensors or targets since deploying more sensors or targets leads to an increase in network load and MC's burden. The high variance can be observed in all settings that demonstrate the dependence on the network topology of the charging algorithms. However, our method dominates INMA, NJNP, and Random in all settings. When $\gls{not:num:sn} = 20$, the DRL-TCC extends the network lifetime to around $36375s$ on average of $1000$ network instances. Meanwhile, the \gls{mc} running with NJNP, INMA, and Random strategy, only conserves the network to around $25600s$, $22027s$, and $3780s$, respectively. Moreover, there are $110$ out of $1000$ network instances at which the MC with DRL-TCC elongates the network to over $2000$ charging actions. While the numbers of NJNP, INMA, and Random are $76$, $60$, and $0$, respectively.

\textbf{Impact of $\kappa$. } We vary the data generation rate $\kappa$ from $0.4$ to $1.0$ and report the comparison in Fig.~\ref{fig:lifetime:1c} and~\ref{fig:lifetime:2c}. Similar to the experiments with sensors and targets, the average lifetime deteriorates as increasing the data generation rate. However, DRL-TCC still shows superior results compared to NJNP and INMA. Specifically, the result of DRL-TCC outperforms $30.57\%$ on average compared to the result given by NJNP and $57.33\%$ better than the result of INMA. The Random strategy continues to show the worst results.

\textbf{Self-organizing capability. } 
To further probe the self-organizing capability of the proposed method, we report the aggregated energy consumption rate and the average number of node failures associated with the aforementioned experiments. The former is the average amount of energy consumed by the sensors in a unit of time and the latter is the average number of exhausted nodes. We report the average over all instances for both metrics in Fig.~\ref{fig:self-org}.

We can observe that the aggregated energy consumption rate of DRL-TCC is slightly lower than others while the number of node failures is higher than INMA and NJNP when increasing the network load. Coupled with the superiority in the performance of prolonging the network's lifetime, it suggests that the DRL-TCC decided to leave \textit{non-critical} nodes to be exhausted to reduce the network load. The transmission topology will be re-organized with a lower network load. This enables the MC to alleviate its burden; it thus elongates the network's lifetime while still ensuring coverage and connectivity. Notice that, despite showing higher performance in the network's lifetime than the Random strategy, the aggregated energy consumption rate of INMA and NJNP has no significant difference compared to Random.

\section{Conclusion}
This paper studied the connected target coverage problem under the settings of the WRSN paradigm. We proposed a novel scheme for scheduling the charging path for the MC to prolong the network lifetime. Our scheme relies on deep reinforcement learning techniques to design a charging policy that takes the network state as the input and outputs the probability of each charging action. Our model is a combination of attention and pointing mechanisms that facilitates the operation of a dynamic network where the number of sensors might be changed due to node failures or deployments. The empirical results show the superiority of our method compared to the existing on-demand algorithms.

\section*{Acknowledgement}
This research is funded by Vietnam National Foundation for Science and Technology Development (NAFOSTED) under grant number 102.01-2019.302.

\bibliographystyle{IEEEtran}
\bibliography{main.bbl}

\begin{thebibliography}{10}
\providecommand{\url}[1]{#1}
\csname url@samestyle\endcsname
\providecommand{\newblock}{\relax}
\providecommand{\bibinfo}[2]{#2}
\providecommand{\BIBentrySTDinterwordspacing}{\spaceskip=0pt\relax}
\providecommand{\BIBentryALTinterwordstretchfactor}{4}
\providecommand{\BIBentryALTinterwordspacing}{\spaceskip=\fontdimen2\font plus
\BIBentryALTinterwordstretchfactor\fontdimen3\font minus
  \fontdimen4\font\relax}
\providecommand{\BIBforeignlanguage}[2]{{%
\expandafter\ifx\csname l@#1\endcsname\relax
\typeout{** WARNING: IEEEtran.bst: No hyphenation pattern has been}%
\typeout{** loaded for the language `#1'. Using the pattern for}%
\typeout{** the default language instead.}%
\else
\language=\csname l@#1\endcsname
\fi
#2}}
\providecommand{\BIBdecl}{\relax}
\BIBdecl

\bibitem{pirbhulal2016novel}
S.~Pirbhulal, H.~Zhang, M.~E. E~Alahi, H.~Ghayvat, S.~C. Mukhopadhyay, Y.-T.
  Zhang, and W.~Wu, ``A novel secure iot-based smart home automation system
  using a wireless sensor network,'' \emph{Sensors}, vol.~17, no.~1, p.~69,
  2016.

\bibitem{kingsy2019comprehensive}
R.~Kingsy~Grace and S.~Manju, ``A comprehensive review of wireless sensor
  networks based air pollution monitoring systems,'' \emph{Wireless Personal
  Communications}, vol. 108, no.~4, pp. 2499--2515, 2019.

\bibitem{alphonsa2016earthquake}
A.~Alphonsa and G.~Ravi, ``Earthquake early warning system by iot using
  wireless sensor networks,'' in \emph{2016 International conference on
  wireless communications, signal processing and networking (WiSPNET)}.\hskip
  1em plus 0.5em minus 0.4em\relax IEEE, 2016, pp. 1201--1205.

\bibitem{sanjeevi2020precision}
P.~Sanjeevi, S.~Prasanna, B.~Siva~Kumar, G.~Gunasekaran, I.~Alagiri, and
  R.~Vijay~Anand, ``Precision agriculture and farming using internet of things
  based on wireless sensor network,'' \emph{Transactions on Emerging
  Telecommunications Technologies}, vol.~31, no.~12, p. e3978, 2020.

\bibitem{abdulkarem2020wireless}
M.~Abdulkarem, K.~Samsudin, F.~Z. Rokhani, and M.~F. A~Rasid, ``Wireless sensor
  network for structural health monitoring: a contemporary review of
  technologies, challenges, and future direction,'' \emph{Structural Health
  Monitoring}, vol.~19, no.~3, pp. 693--735, 2020.

\bibitem{gardavsevic2020emerging}
G.~Garda{\v{s}}evi{\'c}, K.~Katzis, D.~Baji{\'c}, and L.~Berbakov, ``Emerging
  wireless sensor networks and internet of things technologies—foundations of
  smart healthcare,'' \emph{Sensors}, vol.~20, no.~13, p. 3619, 2020.

\bibitem{csaji2017wireless}
B.~C. Cs{\'a}ji, Z.~Kem{\'e}ny, G.~Pedone, A.~Kuti, and J.~V{\'a}ncza,
  ``Wireless multi-sensor networks for smart cities: A prototype system with
  statistical data analysis,'' \emph{IEEE Sensors Journal}, vol.~17, no.~23,
  pp. 7667--7676, 2017.

\bibitem{tripathi2018coverage}
A.~Tripathi, H.~P. Gupta, T.~Dutta, R.~Mishra, K.~Shukla, and S.~Jit,
  ``Coverage and connectivity in wsns: A survey, research issues and
  challenges,'' \emph{IEEE Access}, vol.~6, pp. 26\,971--26\,992, 2018.

\bibitem{zhao2008lifetime}
Q.~Zhao and M.~Gurusamy, ``Lifetime maximization for connected target coverage
  in wireless sensor networks,'' \emph{IEEE/ACM Transactions on Networking},
  2008.

\bibitem{akyildiz2002wireless}
I.~F. Akyildiz, W.~Su, Y.~Sankarasubramaniam, and E.~Cayirci, ``Wireless sensor
  networks: a survey,'' \emph{Computer networks}, vol.~38, no.~4, pp. 393--422,
  2002.

\bibitem{goyal2019data}
N.~Goyal, M.~Dave, and A.~K. Verma, ``Data aggregation in underwater wireless
  sensor network: Recent approaches and issues,'' \emph{Journal of King Saud
  University-Computer and Information Sciences}, vol.~31, no.~3, pp. 275--286,
  2019.

\bibitem{haimour2019energy}
J.~Haimour and O.~Abu-Sharkh, ``Energy efficient sleep/wake-up techniques for
  iot: A survey,'' in \emph{2019 IEEE Jordan International Joint Conference on
  Electrical Engineering and Information Technology (JEEIT)}.\hskip 1em plus
  0.5em minus 0.4em\relax IEEE, 2019, pp. 478--484.

\bibitem{raj2019qos}
J.~S. Raj, A.~Basar \emph{et~al.}, ``Qos optimization of energy efficient
  routing in iot wireless sensor networks,'' \emph{Journal of ISMAC}, vol.~1,
  no.~01, pp. 12--23, 2019.

\bibitem{adu2018energy}
K.~S. Adu-Manu, N.~Adam, C.~Tapparello, H.~Ayatollahi, and W.~Heinzelman,
  ``Energy-harvesting wireless sensor networks (eh-wsns) a review,'' \emph{ACM
  Transactions on Sensor Networks (TOSN)}, vol.~14, no.~2, pp. 1--50, 2018.

\bibitem{kurs2007wireless}
A.~Kurs, A.~Karalis, R.~Moffatt, J.~D. Joannopoulos, P.~Fisher, and
  M.~Solja{\v{c}}i{\'c}, ``Wireless power transfer via strongly coupled
  magnetic resonances,'' \emph{science}, 2007.

\bibitem{lu2015wireless}
X.~Lu, P.~Wang, D.~Niyato, D.~I. Kim, and Z.~Han, ``Wireless charging
  technologies: Fundamentals, standards, and network applications,'' \emph{IEEE
  communications surveys \& tutorials}, vol.~18, no.~2, pp. 1413--1452, 2015.

\bibitem{he2012energy}
S.~He, J.~Chen, F.~Jiang, D.~K. Yau, G.~Xing, and Y.~Sun, ``Energy provisioning
  in wireless rechargeable sensor networks,'' \emph{IEEE transactions on mobile
  computing}, 2012.

\bibitem{lyu2019periodic}
Z.~Lyu, Z.~Wei, J.~Pan, H.~Chen, C.~Xia, J.~Han, and L.~Shi, ``Periodic
  charging planning for a mobile wce in wireless rechargeable sensor networks
  based on hybrid pso and ga algorithm,'' \emph{Applied Soft Computing}, 2019.

\bibitem{jiang2017joint}
G.~Jiang, S.-K. Lam, Y.~Sun, L.~Tu, and J.~Wu, ``Joint charging tour planning
  and depot positioning for wireless sensor networks using mobile chargers,''
  \emph{IEEE/ACM Transactions on Networking}, 2017.

\bibitem{ma2018charging}
Y.~Ma, W.~Liang, and W.~Xu, ``Charging utility maximization in wireless
  rechargeable sensor networks by charging multiple sensors simultaneously,''
  \emph{IEEE/ACM Transactions on Networking}, vol.~26, no.~4, 2018.

\bibitem{xu2019minimizing}
W.~Xu, W.~Liang, H.~Kan, Y.~Xu, and X.~Zhang, ``Minimizing the longest charge
  delay of multiple mobile chargers for wireless rechargeable sensor networks
  by charging multiple sensors simultaneously,'' in \emph{2019 IEEE 39th
  International Conference on Distributed Computing Systems (ICDCS)}.\hskip 1em
  plus 0.5em minus 0.4em\relax IEEE, 2019, pp. 881--890.

\bibitem{he2013demand}
L.~He, Y.~Gu, J.~Pan, and T.~Zhu, ``On-demand charging in wireless sensor
  networks: Theories and applications,'' in \emph{2013 IEEE 10th International
  Conference on Mobile Ad-hoc and Sensor Systems}.\hskip 1em plus 0.5em minus
  0.4em\relax IEEE, 2013, pp. 28--36.

\bibitem{fu2015esync}
L.~Fu, L.~He, P.~Cheng, Y.~Gu, J.~Pan, and J.~Chen, ``Esync: Energy
  synchronized mobile charging in rechargeable wireless sensor networks,''
  \emph{IEEE Transactions on vehicular technology}, vol.~65, no.~9, pp.
  7415--7431, 2015.

\bibitem{lin2017tsca}
C.~Lin, J.~Zhou, C.~Guo, H.~Song, G.~Wu, and M.~S. Obaidat, ``Tsca: A
  temporal-spatial real-time charging scheduling algorithm for on-demand
  architecture in wireless rechargeable sensor networks,'' \emph{IEEE
  Transactions on Mobile Computing}, vol.~17, no.~1, pp. 211--224, 2017.

\bibitem{lin2019double}
C.~Lin, Y.~Sun, K.~Wang, Z.~Chen, B.~Xu, and G.~Wu, ``Double warning thresholds
  for preemptive charging scheduling in wireless rechargeable sensor
  networks,'' \emph{Computer Networks}, vol. 148, pp. 72--87, 2019.

\bibitem{zhu2018adaptive}
J.~Zhu, Y.~Feng, M.~Liu, G.~Chen, and Y.~Huang, ``Adaptive online mobile
  charging for node failure avoidance in wireless rechargeable sensor
  networks,'' \emph{Computer Communications}, vol. 126, pp. 28--37, 2018.

\bibitem{kaswan2018efficient}
A.~Kaswan, A.~Tomar, and P.~K. Jana, ``An efficient scheduling scheme for
  mobile charger in on-demand wireless rechargeable sensor networks,''
  \emph{Journal of Network and Computer Applications}, vol. 114, pp. 123--134,
  2018.

\bibitem{la2020q}
V.~Q. La, P.~Le~Nguyen, T.-H. Nguyen, K.~Nguyen \emph{et~al.},
  ``Q-learning-based, optimized on-demand charging algorithm in {WRSN},'' in
  \emph{2020 IEEE 19th International Symposium on Network Computing and
  Applications (NCA)}.\hskip 1em plus 0.5em minus 0.4em\relax IEEE, 2020, pp.
  1--8.

\bibitem{cao2021deep}
X.~Cao, W.~Xu, X.~Liu, J.~Peng, and T.~Liu, ``A deep reinforcement
  learning-based on-demand charging algorithm for wireless rechargeable sensor
  networks,'' \emph{Ad Hoc Networks}, vol. 110, p. 102278, 2021.

\bibitem{Gawade2016}
R.~D. Gawade and S.~L. Nalbalwar, ``{A centralized energy efficient distance
  based routing protocol for wireless sensor networks},'' \emph{Journal of
  Sensors}, vol. 2016, 2016.

\bibitem{Wu2013}
Y.~Wu and W.~Liu, ``{Routing protocol based on genetic algorithm for energy
  harvesting-wireless sensor networks},'' \emph{IET}, vol.~3, no.~2, pp.
  112--118, 2013.

\bibitem{hottung2019neural}
A.~Hottung and K.~Tierney, ``Neural large neighborhood search for the
  capacitated vehicle routing problem,'' \emph{arXiv preprint
  arXiv:1911.09539}, 2019.

\bibitem{vinyals2015pointer}
O.~Vinyals, M.~Fortunato, and N.~Jaitly, ``Pointer networks,'' \emph{arXiv
  preprint arXiv:1506.03134}, 2015.

\bibitem{williams1992simple}
R.~J. Williams, ``Simple statistical gradient-following algorithms for
  connectionist reinforcement learning,'' \emph{Machine Learning}, vol.~8, no.
  3-4, pp. 229--256, 1992.

\bibitem{schulman2015high}
J.~Schulman, P.~Moritz, S.~Levine, M.~Jordan, and P.~Abbeel, ``High-dimensional
  continuous control using generalized advantage estimation,'' \emph{arXiv
  preprint arXiv:1506.02438}, 2015.

\bibitem{mnih2016asynchronous}
V.~Mnih, A.~P. Badia, M.~Mirza, A.~Graves, T.~Lillicrap, T.~Harley, D.~Silver,
  and K.~Kavukcuoglu, ``Asynchronous methods for deep reinforcement learning,''
  in \emph{International Conference on Machine Learning}.\hskip 1em plus 0.5em
  minus 0.4em\relax PMLR, 2016.

\bibitem{kingma2014adam}
D.~P. Kingma and J.~Ba, ``Adam: A method for stochastic optimization,''
  \emph{arXiv preprint arXiv:1412.6980}, 2014.

\end{thebibliography}

\end{document}